\documentclass[11pt,a4paper,logo]{personalpaper}
\graphicspath{{figs/}}

\makeatletter
\DeclareRobustCommand\onedot{\futurelet\@let@token\@onedot}
\def\@onedot{\ifx\@let@token.\else.\null\fi\xspace}

\makeatother

\usepackage{hyperref}
\usepackage{url}

\usepackage{amsmath}
\usepackage{amssymb}
\usepackage{mathtools}
\usepackage{amsthm}

\usepackage[table]{xcolor}
\usepackage{multirow}
\usepackage{makecell}
\usepackage{pifont}       
\usepackage{bbding}      

\usepackage{subcaption}
\usepackage{arydshln}
\usepackage{tabularx}
\usepackage{mdframed}
\usepackage[most]{tcolorbox}
\usepackage{arydshln}
\usepackage{adjustbox}
\usepackage{wrapfig}
\usepackage{titletoc}

\usepackage{booktabs,array,colortbl,graphicx}
\usepackage{algorithm}
\usepackage{algpseudocode}
\usepackage{caption}
\usepackage{wrapfig}

\definecolor{table-blue}{RGB}{173, 216, 230}
\definecolor{row-highlight}{RGB}{220, 230, 242}
\definecolor{qwen-header}{RGB}{235, 240, 248}
\definecolor{llama-header}{RGB}{248, 244, 235}
\definecolor{section-gray}{RGB}{245, 245, 245}
\definecolor{group-header}{RGB}{230, 235, 245}

\definecolor{zhz_gray}{rgb}{0.8,0.8,0.8}
\definecolor{darkgreen}{rgb}{0.0, 0.5, 0.0} % Custom dark green color
\definecolor{darkred}{rgb}{0.5, 0.0, 0.0}   % Custom dark red color

\definecolor{row-alt}{RGB}{248, 250, 253}      % very light cool tint
\definecolor{hl-llama}{RGB}{234, 242, 255}     % soft blue
\definecolor{hl-qwen}{RGB}{236, 250, 240}      % soft green

\definecolor{tableheader}{HTML}{F2F3F5}
\definecolor{oursblue}{HTML}{EEF5FF}
\definecolor{panelgray}{HTML}{E8EAED}

\definecolor{datasetbg}{HTML}{ECEFF2}

\definecolor{f1header}{HTML}{C6DDF1}
\definecolor{ljheader}{HTML}{C7E4D0}
\definecolor{costheader}{HTML}{F1D5A7}
\definecolor{latheader}{HTML}{DAC7E7}

\definecolor{f1best}{HTML}{D2E4F4}
\definecolor{ljbest}{HTML}{D2E9D8}
\definecolor{costbest}{HTML}{F4DCB4}
\definecolor{latbest}{HTML}{E2D3EC}

\definecolor{f1second}{HTML}{EAF3FA}
\definecolor{ljsecond}{HTML}{ECF6EF}
\definecolor{costsecond}{HTML}{FBF0DE}
\definecolor{latsecond}{HTML}{F2EAF6}

\usepackage{tikz}

\newcommand{\blacknum}[1]{%
  \tikz[baseline=(char.base)]{
    \node[
      shape=circle,
      fill=black,
      text=white,
      inner sep=0.8pt,
      font=\scriptsize\bfseries
    ] (char) {#1};
  }%
}

\definecolor{retrievebg}{HTML}{DCEAF7}
\definecolor{retrievetext}{HTML}{315F86}
\definecolor{curatebg}{HTML}{F6E4C5}
\definecolor{curatetext}{HTML}{8A5A16}

\newcommand{\Retrieve}{%
  \begingroup
  \setlength{\fboxsep}{1.0pt}%
  \colorbox{retrievebg}{\textcolor{retrievetext}{\textsc{\textbf{Retrieve}}}}%
  \endgroup
}

\newcommand{\Curate}{%
  \begingroup
  \setlength{\fboxsep}{1.0pt}%
  \colorbox{curatebg}{\textcolor{curatetext}{\textsc{\textbf{Curate}}}}%
  \endgroup
}

\definecolor{mempilotblue}{HTML}{2563EB}
\definecolor{mempilotmid}{HTML}{7C3AED}
\definecolor{mempilotorange}{HTML}{D97706}

\newcommand{\MemPilotName}{%
  \mbox{\textbf{%
    \textcolor{mempilotblue}{M}%
    \textcolor{mempilotblue!67!mempilotmid}{e}%
    \textcolor{mempilotblue!33!mempilotmid}{m}%
    \textcolor{mempilotmid}{P}%
    \textcolor{mempilotmid!75!mempilotorange}{i}%
    \textcolor{mempilotmid!50!mempilotorange}{l}%
    \textcolor{mempilotmid!25!mempilotorange}{o}%
    \textcolor{mempilotorange}{t}%
  }}%
}

\RequirePackage{listings}
\RequirePackage{cleveref}
\RequirePackage{soul}
\tcbuselibrary{listings}

\newcounter{ppalgorithm}

\theoremstyle{plain}

\theoremstyle{definition}

\theoremstyle{remark}

\tcolorboxenvironment{theorem}{
  colback=ppsoftblue,
  colframe=ppnavy!70,
  arc=1mm,
  boxrule=.5pt,
  top=.7mm,
  bottom=.7mm,
  left=1mm,
  right=1mm,
  before skip=6pt,
  after skip=6pt
}

\tcolorboxenvironment{proposition}{
  colback=ppsoftblue,
  colframe=ppnavy!60,
  arc=1mm,
  boxrule=.5pt,
  top=.7mm,
  bottom=.7mm,
  left=1mm,
  right=1mm,
  before skip=6pt,
  after skip=6pt
}

\tcolorboxenvironment{definition}{
  colback=ppabsbg,
  colframe=ppaccent!70!black,
  arc=1mm,
  boxrule=.5pt,
  top=.7mm,
  bottom=.7mm,
  left=1mm,
  right=1mm,
  before skip=6pt,
  after skip=6pt
}

\tcolorboxenvironment{remark}{
  colback=ppsoftgray,
  colframe=ppmuted!70,
  arc=1mm,
  boxrule=.45pt,
  top=.7mm,
  bottom=.7mm,
  left=1mm,
  right=1mm,
  before skip=6pt,
  after skip=6pt
}

\tcolorboxenvironment{proof}{
  enhanced,
  breakable,
  colback=ppsoftblue!35!white,
  colframe=ppaccent!65!ppnavy,
  arc=1mm,
  boxrule=.4pt,
  top=.7mm,
  bottom=.7mm,
  left=1mm,
  right=1mm,
  before skip=6pt,
  after skip=8pt
}

\newtcolorbox{takeawaybox}[1][Takeaway]{
  enhanced,
  breakable,
  title=#1,
  colback=ppabsbg,
  colframe=ppnavy,
  fonttitle=\bfseries\small,
  coltitle=white,
  colbacktitle=ppnavy,
  attach boxed title to top left={xshift=2.5mm,yshift=-2.5mm},
  boxed title style={rounded corners, size=small, colframe=ppnavy, colback=ppnavy},
  arc=2mm,
  width=\linewidth
}

\newtcolorbox{highlightbox}[1][Highlight]{
  enhanced,
  breakable,
  title=#1,
  colback=ppsoftblue,
  colframe=ppaccent!80!black,
  fonttitle=\bfseries\small,
  coltitle=white,
  colbacktitle=ppaccent!80!black,
  attach boxed title to top left={xshift=2.5mm,yshift=-2.5mm},
  boxed title style={rounded corners, size=small, colframe=ppaccent!80!black, colback=ppaccent!80!black},
  arc=2mm,
  width=\linewidth
}

\definecolor{pppromptamber}{HTML}{B7791F}
\definecolor{pppromptblue}{HTML}{315F9D}
\definecolor{pppromptgreen}{HTML}{2F7D6D}
\definecolor{pppromptpurple}{HTML}{6B5CA5}
\definecolor{pppromptred}{HTML}{A23A34}
\definecolor{pppromptgray}{HTML}{69727D}
\definecolor{ppprompthl}{HTML}{FFE8A3}

\tcbset{
  pppromptbase/.style={
    enhanced,
    breakable,
    fonttitle=\bfseries\small,
    coltitle=white,
    attach boxed title to top left={xshift=2.5mm,yshift=-2.5mm},
    width=\linewidth,
    arc=2mm,
    boxrule=.55pt,
    top=2.8mm,
    bottom=1.2mm,
    left=1.2mm,
    right=1.2mm,
    before skip=8pt,
    after skip=8pt,
    fontupper=\small
  },
  promptbox/.style={
    pppromptbase,
    colback=pppromptamber!8,
    colframe=pppromptamber!80!black,
    colbacktitle=pppromptamber!80!black,
    boxed title style={rounded corners, size=small, boxrule=.4pt, colframe=pppromptamber!80!black, colback=pppromptamber!80!black}
  },
  promptboxblue/.style={
    pppromptbase,
    colback=pppromptblue!7,
    colframe=pppromptblue!80!black,
    colbacktitle=pppromptblue!80!black,
    boxed title style={rounded corners, size=small, boxrule=.4pt, colframe=pppromptblue!80!black, colback=pppromptblue!80!black}
  },
  promptboxgreen/.style={
    pppromptbase,
    colback=pppromptgreen!8,
    colframe=pppromptgreen!75!black,
    colbacktitle=pppromptgreen!75!black,
    boxed title style={rounded corners, size=small, boxrule=.4pt, colframe=pppromptgreen!75!black, colback=pppromptgreen!75!black}
  },
  promptboxpurple/.style={
    pppromptbase,
    colback=pppromptpurple!8,
    colframe=pppromptpurple!78!black,
    colbacktitle=pppromptpurple!78!black,
    boxed title style={rounded corners, size=small, boxrule=.4pt, colframe=pppromptpurple!78!black, colback=pppromptpurple!78!black}
  },
  promptboxred/.style={
    pppromptbase,
    colback=pppromptred!6,
    colframe=pppromptred!85!black,
    colbacktitle=pppromptred!85!black,
    boxed title style={rounded corners, size=small, boxrule=.4pt, colframe=pppromptred!85!black, colback=pppromptred!85!black}
  },
  promptboxgray/.style={
    pppromptbase,
    colback=ppsoftgray,
    colframe=pppromptgray!78!black,
    colbacktitle=pppromptgray!78!black,
    boxed title style={rounded corners, size=small, boxrule=.4pt, colframe=pppromptgray!78!black, colback=pppromptgray!78!black}
  }
}

\newtcblisting{promptlisting}[2][promptbox]{
  #1,
  title={#2},
  listing only,
  listing engine=listings,
  listing options={
    basicstyle=\ttfamily\scriptsize,
    breaklines=true,
    columns=fullflexible,
    keepspaces=true,
    showstringspaces=false
  }
}

\lstdefinestyle{python}{
  language=Python,
  basicstyle=\ttfamily\small,
  keywordstyle=\color{pplink}\bfseries,
  stringstyle=\color{ppgood!80!black},
  commentstyle=\color{ppmuted}\itshape,
  showstringspaces=false,
  numbers=left,
  numberstyle=\tiny\color{ppmuted},
  stepnumber=1,
  numbersep=6pt,
  frame=single,
  rulecolor=\color{ppnavy!40},
  backgroundcolor=\color{ppsoftgray},
  breaklines=true,
  columns=fullflexible,
  keepspaces=true
}

\crefname{theorem}{Theorem}{Theorems}
\crefname{proposition}{Proposition}{Propositions}
\crefname{lemma}{Lemma}{Lemmas}
\crefname{corollary}{Corollary}{Corollaries}
\crefname{definition}{Definition}{Definitions}
\crefname{assumption}{Assumption}{Assumptions}
\crefname{remark}{Remark}{Remarks}
\crefname{ppalgorithm}{Algorithm}{Algorithms}
\Crefname{ppalgorithm}{Algorithm}{Algorithms}
\crefname{algorithm}{Algorithm}{Algorithms}
\Crefname{algorithm}{Algorithm}{Algorithms}
\crefname{lstlisting}{Listing}{Listings}
\Crefname{lstlisting}{Listing}{Listings}

\newcommand{\projectlinktext}{Project}
\newcommand{\codelinktext}{Code}
\newcommand{\modellinktext}{Models}

\newcommand{\setprojectlink}[2]{%
  \projecturl{#1}%
  \renewcommand{\projectlinktext}{#2}%
}

\newcommand{\setcodelink}[2]{%
  \codeurl{#1}%
  \renewcommand{\codelinktext}{#2}%
}

\newcommand{\setmodellink}[2]{%
  \modelurl{#1}%
  \renewcommand{\modellinktext}{#2}%
}

\makeatletter

\renewcommand{\pp@contactline}{}

\newcommand{\pp@plainfootnotetext}[1]{\noindent #1}

\g@addto@macro\maketitle{%
  \if\relax\the\correspondingauthor\relax
  \else
    \begingroup
      \let\@makefntext\pp@plainfootnotetext
      \footnotetext[0]{%
        \textit{Corresponding author(s):}
        \href{mailto:\the\correspondingauthor}{\the\correspondingauthor}%
      }%
    \endgroup
  \fi
}

\renewcommand{\pp@linkline}{%
  \begingroup
  \centering
  \if\relax\the\projecturl\relax\else
    {\linkfont \projectlinkicon\
      \href{\the\projecturl}{\projectlinktext}}\quad\quad\quad\quad
  \fi
  \if\relax\the\codeurl\relax\else
    {\linkfont \faGithub\
      \href{\the\codeurl}{\codelinktext}}\quad\quad\quad\quad
  \fi
  \if\relax\the\modelurl\relax\else
    {\linkfont \modellinkicon\
      \href{\the\modelurl}{\modellinktext}}%
  \fi
  \par
  \endgroup
}

\makeatother

\setprojectlink{https://example.com/project}{Project Website}
\setcodelink{https://github.com/ViktorAxelsen/MemPilot}{GitHub Repository}
\setmodellink{https://example.com/models}{Model Checkpoints}

\correspondingauthor{wazhz14@gmail.com}

\setheadertext{GLINT LAB}
\setheadertitle{\MemPilotName: Orchestrating On-Demand Multimodal Memory Curation for LLM Agents}
\setheaderlogos{%
  \headerlogo[25pt]{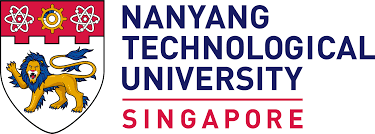}\headerlogosep
  \headerlogo[21pt]{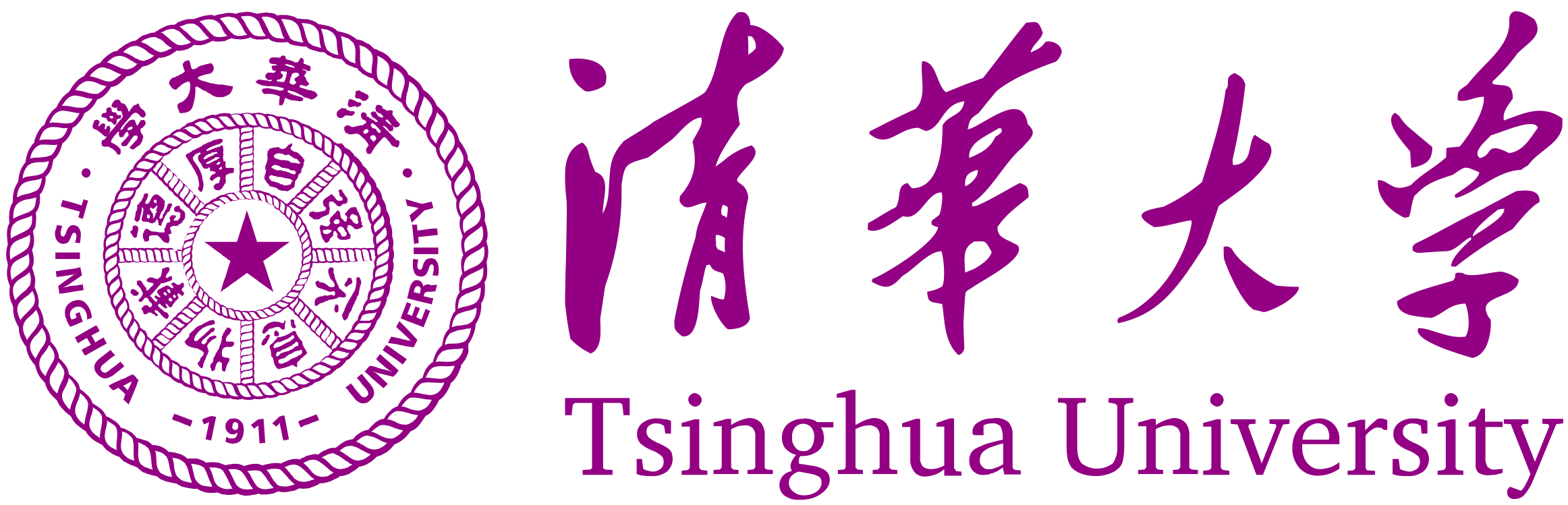}\headerlogosep
  \headerlogo[18pt]{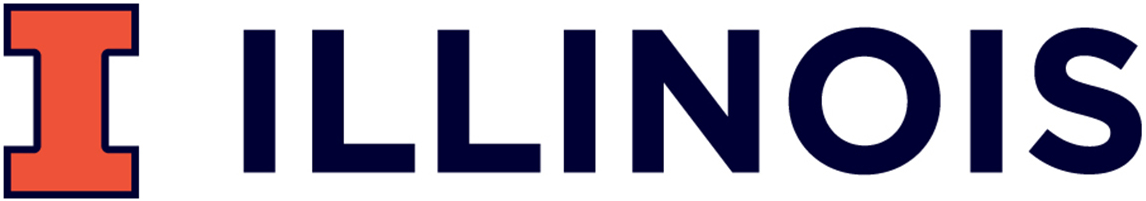}%
}
\setheaderlogoraise{10pt}
\setprojectlink{https://viktoraxelsen.github.io/MemPilot/}{Project Website}
\setcodelink{https://github.com/ViktorAxelsen/MemPilot}{MemPilot}
\setmodellink{https://hf.co/collections/XaiverZ/mempilot}{MemPilot Collection}

\correspondingauthor{wazhz14@gmail.com}

\title{\MemPilotName: Orchestrating On-Demand Multimodal Memory Curation for LLM Agents}

\author[1]{Haozhen Zhang}
\author[2]{Haodong Yue}
\author[1]{Quanyu Long}
\author[1]{Jianzhu Bao}
\author[1]{Qingyuan Liu}
\author[3]{Tao Feng}
\author[1]{Bohan Liu}
\author[1]{Weida Liang}
\author[1]{Wenya Wang}

\affil[1]{Nanyang Technological University}
\affil[2]{Tsinghua University}
\affil[3]{University of Illinois Urbana-Champaign}

\begin{abstract}

Memory has become integral to the LLM agent ecosystem, supporting information retention and reuse across interactions.
However, most existing agent memory systems construct memory in a query-agnostic manner, which can incur unnecessary preprocessing cost and discard details that later prove essential.
Recent studies have begun shifting memory processing toward runtime adaptation, but typically specialize in particular operations or fixed processing schemes, leaving flexible control over performance, cost, and latency largely underexplored.
To address this challenge, we present \textbf{MemPilot}, a flexible framework that orchestrates on-demand memory curation under different performance--cost--latency preferences.
Specifically, we optimize a multi-step LLM policy via reinforcement learning to iteratively choose between retrieving from query-agnostic memory and delegating query-specific curation of raw multimodal history to heterogeneous LLMs and VLMs.
The policy jointly controls evidence amount, curation instructions, model selection, and visual access, enabling fine-grained allocation of runtime computation.
To optimize this policy under competing objectives, we adapt objective-wise advantage decoupling by separately estimating each objective's advantage before aggregation.
Moreover, we introduce prefix-based marginal utility estimation for fine-grained credit assignment across multi-step rollouts.
Experiments on five multimodal agent-memory benchmarks demonstrate favorable performance--cost--latency trade-offs across optimization preferences, with preference sweeps yielding broader frontiers than existing trade-off-aware baselines.

\end{abstract}

\begin{document}
\maketitle

\section{Introduction}

As large language model (LLM) agents evolve toward persistent, long-horizon interaction, memory becomes increasingly central to preserving and reusing past experience~\citep{hu2025memory, huang2026rethinking}.
However, most existing agent memory systems construct memory from interaction histories in a query-agnostic manner, and relying solely on such pre-query construction can be wasteful and lossy, as it cannot reliably anticipate future query needs~\citep{xu2025mem, fang2025lightmem, liu2026simplemem}.
To mitigate this limitation, a growing line of work moves memory adaptation to runtime through query-time evidence distillation, on-demand memory construction, and adaptive memory strategies~\citep{yan2025general, zhang2026adamem}.
Despite this progress, flexible orchestration of runtime memory adaptation remains underexplored, particularly when adapting runtime computation to query needs and multimodal evidence while balancing performance, cost, and latency~\citep{zhang2026learning}.
This raises a key challenge: \emph{how can runtime multimodal memory curation be adaptively orchestrated with controllable performance--cost--latency trade-offs?}

Unfortunately, addressing this challenge is non-trivial due to three key limitations of existing approaches to runtime memory adaptation.
\emph{\blacknum{1} Runtime memory control remains structurally rigid}: existing methods typically rely on hand-designed operations or fixed processing pipelines, with efficiency control hard-wired into predefined modules, budget tiers, or heuristics~\citep{yin2026defermem, zhang2026learning, yu2026lazymem}, leaving few degrees of freedom for query-specific adaptation and making fine-grained trade-off optimization difficult.
\emph{\blacknum{2} Trade-off optimization remains incomplete}: existing approaches often target only a subset of efficiency objectives, such as token usage or monetary cost, while user-facing latency is rarely treated as an explicit optimization objective~\citep{liu2026omni, zhang2026adamem}.
\emph{\blacknum{3} Existing methods largely assume a fixed processing backend for runtime memory adaptation}, overlooking substantial variation in curation quality, cost, and latency across heterogeneous models and limiting adaptation to query difficulty and evidence modality, thereby constraining the achievable trade-off space~\citep{yu2026lazymem, yan2025general}.
Collectively, these limitations call for a more flexible framework that can orchestrate runtime memory operations and processing backends according to both query needs and optimization preferences.

Building on these observations, in this work we introduce \textbf{MemPilot}, a flexible framework for orchestrating on-demand memory curation under different performance--cost--latency preferences.
At a high level, MemPilot follows a simple design principle: retain query-agnostic memory, compatible with off-the-shelf agent memory systems, for efficient access, while preserving raw interaction history for query-specific runtime curation.
Specifically, at inference time, a multi-step LLM policy constructs a query-adaptive orchestration trajectory by iteratively choosing between retrieving from the query-agnostic memory bank and delegating curation of the raw history to heterogeneous LLMs and vision-language models (VLMs), whose distinct capabilities naturally offer different quality, cost, and latency profiles.
Rather than relying on hand-crafted pipelines or discrete budget tiers, the policy adaptively determines how much evidence to process, how it should be curated, which model to invoke, and whether visual evidences are needed, exposing fine-grained controls over runtime computation trade-offs.
We optimize this policy with reinforcement learning under different performance, cost, and latency preferences. To better handle the resulting multi-objective optimization, we adapt objective-wise advantage decoupling by estimating each objective's advantage separately before weighted aggregation.
We further introduce prefix-based marginal utility estimation, which estimates the marginal contribution of individual curation steps for fine-grained credit assignment.

Experiments on five multimodal agent-memory benchmarks show that MemPilot achieves favorable performance--cost--latency trade-offs across different optimization preferences. Fine-grained preference sweeps further reveal broader trade-off frontiers than existing trade-off-aware baselines, while behavioral analyses show that MemPilot adaptively reallocates runtime computation as optimization preferences change.
\textbf{Our contributions are summarized as follows:}
\begin{itemize}
\item We introduce \textbf{MemPilot}, a systematic framework for preference-aware runtime memory orchestration that jointly controls memory access, query-specific curation, heterogeneous model delegation, and visual evidence usage. By unifying these decisions within a multi-step policy, MemPilot enables fine-grained performance--cost--latency control without relying on fixed processing pipelines, predefined budget tiers, or a fixed processing backend.

\item To optimize the orchestration policy under competing objectives, we adapt objective-wise advantage decoupling and introduce prefix-based marginal utility estimation for fine-grained credit assignment in multi-step rollouts, enabling stable and effective optimization.

\item Experiments on five multimodal agent-memory benchmarks demonstrate favorable performance--cost--latency frontiers across preferences, with analyses revealing shifts in model selection, visual access, and computation allocation as preferences change.
\end{itemize}

\section{Related Work}

\subsection{Agent Memory and Runtime Adaptation}

LLM agents increasingly rely on memory mechanisms to preserve and reuse information from long-term interaction histories~\citep{zhong2024memorybank,packer2023memgpt}. 
Most existing text-centric systems construct memory before future queries are known, using either hand-designed mechanisms for extraction and organization~\citep{chhikara2025mem0,xu2025mem,kang2025memory,fang2025lightmem,liu2026simplemem} or learned policies for memory construction, updating, and retrieval~\citep{yan2025memory,wang2025mem,zhang2026memskill,yu2026agentic,yue2026mem}. 
Beyond text-centric memory, recent work has extended agent memory to multimodal interactions, maintaining structured representations over textual, visual, and other sensory experiences~\citep{feng2026m2a,wang2025mirix,long2026seeing,liu2025memverse,liu2026omni}. 
Despite increasingly adaptive and multimodal memory designs, they still construct memory in a largely query-agnostic manner.

A growing line of work instead shifts memory processing toward runtime adaptation. 
GAM~\citep{yan2025general} performs just-in-time research over stored memory and raw history, while BudgetMem~\citep{zhang2026learning} learns query-aware runtime extraction under different computation budgets. 
MemGen~\citep{zhang2025memgen} and AdaMEM~\citep{zhang2026adamem} dynamically generate or refresh memory guidance according to evolving reasoning states, whereas MemRL~\citep{zhang2026memrl} adapts the utility of stored experiences from interaction feedback. 
More recent work further explores query-time evidence distillation and deferred memory construction~\citep{yin2026defermem,yu2026lazymem}. 
However, these approaches typically specialize in a particular form of runtime adaptation or follow a predefined processing pattern. 
MemPilot instead learns to orchestrate heterogeneous multimodal memory curation at runtime under controllable performance--cost--latency preferences.

\subsection{LLM Routing}

LLM routing aims to match different queries with suitable models based on their heterogeneous capabilities and costs.
Early approaches route queries through model cascades or select models based on estimated query difficulty~\citep{chen2023frugalgpt,ding2024hybrid}. 
More recent work improves routing through preference learning~\citep{ong2025routellm}, contrastive objectives~\citep{chen2024routerdc}, and graph-based modeling~\citep{feng2024graphrouter}, while further extending routing to multi-agent settings~\citep{yue2025masrouter} and unified routing frameworks~\citep{feng2026llmrouter}.
Moving beyond one-shot routing, several methods employ reinforcement learning to enable multi-step model selection and orchestration~\citep{zhang2025router,qian2025xrouter,feng2026graphplanner}.
These advances highlight the potential of LLM routing as a general paradigm for adaptively allocating heterogeneous model computation according to task demands and resource constraints.

\section{Methodology}

In this section, we first formulate on-demand memory orchestration as a multi-step decision process (Section~\ref{sec:formulation}). We then instantiate the factorized curation and delegation decisions that enable query-adaptive runtime control (Section~\ref{sec:orchestration}). Finally, we introduce resource-aware reward modeling for optimizing quality--cost--latency trade-offs with fine-grained credit assignment (Section~\ref{sec:reward}).

\begin{figure*}[t]
	\centering
	\includegraphics[width=1.0\linewidth]{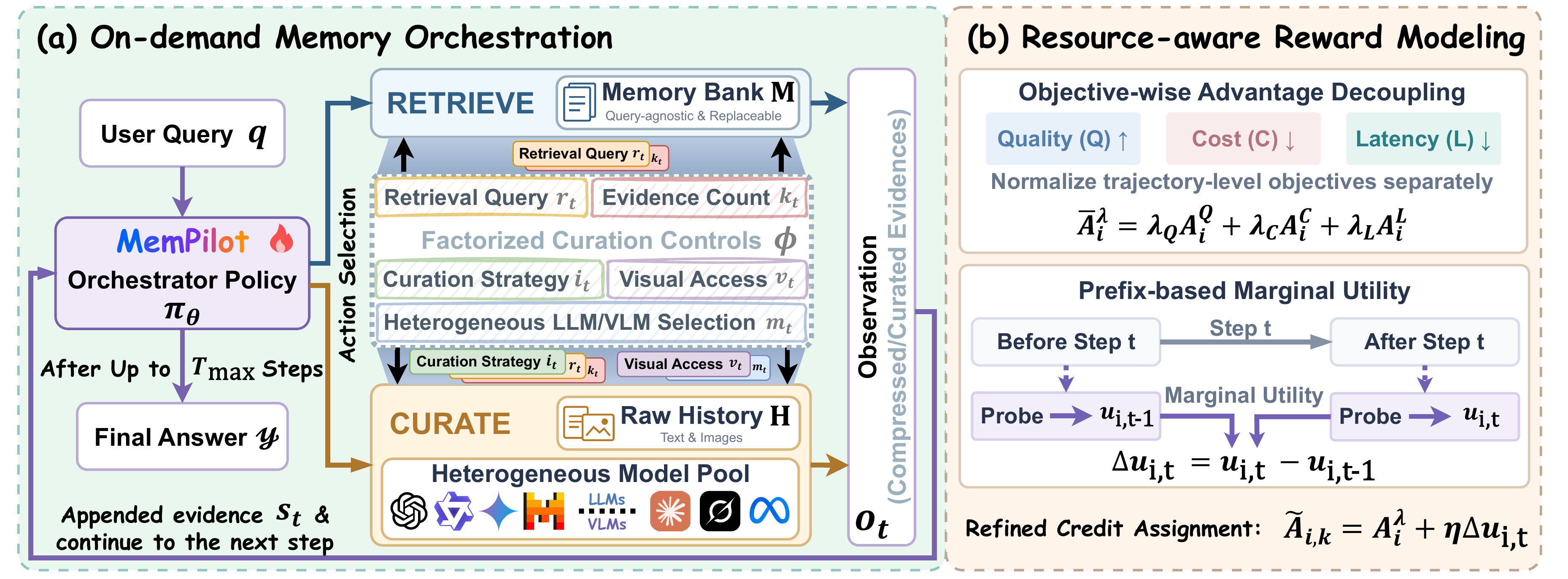}
        \caption{\textbf{Overview of MemPilot.}
\textbf{(a)} The orchestrator iteratively retrieves compressed memories or delegates query-specific curation of raw multimodal histories to heterogeneous LLMs/VLMs. Factorized controls govern evidence retrieval, curation strategy, visual access, and model selection, with accumulated evidence guiding subsequent decisions and final answering.
\textbf{(b)} Objective-wise advantage decoupling enables controllable trade-offs among quality, cost, and latency, while prefix-based answer probing estimates each step's marginal utility for fine-grained credit assignment.}
\label{fig:method}
\end{figure*}

\subsection{On-Demand Memory Orchestration as a Multi-Step Decision Process}
\label{sec:formulation}

\textbf{Dual-view memory representation.}
To support flexible runtime adaptation while retaining efficient access to query-agnostic memory, MemPilot maintains two complementary views of the interaction history: a query-agnostic memory bank $\mathcal{M}$ and the raw multimodal history $\mathcal{H}$. In this work, $\mathcal{M}$ can be initialized with memories constructed by any off-the-shelf agent memory system~\citep{xu2025mem, fang2025lightmem, liu2026simplemem} or a simpler memory construction scheme, keeping memory construction orthogonal to MemPilot's runtime orchestration and ensuring broad compatibility with existing memory systems. Meanwhile, $\mathcal{H}$ is segmented into fixed-length token chunks while preserving the original multimodal context within each chunk. These two views serve complementary roles: the former enables efficient memory access, whereas the latter retains fine-grained information that may be absent from the query-agnostic memory representation.

\textbf{Sequential decision formulation.}
As shown in Figure~\ref{fig:method}, given a query $q$, MemPilot formulates on-demand memory curation as a multi-step decision process governed by an orchestrator policy $\pi_\theta$. 
At step $t$, the orchestrator conditions on the query and previously acquired evidence, $s_t=(q,a_{<t},o_{<t})$, and produces a structured memory action:
\begin{equation}
    a_t=(z_t,\phi_t)\sim\pi_\theta(\cdot\mid s_t),
\end{equation}
where $z_t$ denotes the action type and $\phi_t$ denotes its associated action arguments (for simplicity, we show one action per step, and multiple actions may be issued and run in parallel). Specifically,
\begin{equation}
    z_t\in\{\text{\Retrieve},\,\text{\Curate}\}.
\end{equation}
Here, \Retrieve{} directly accesses the query-agnostic memory bank $\mathcal{M}$, while \Curate{} performs query-specific runtime curation over the raw history $\mathcal{H}$. We abstract the resulting observation as:
\begin{equation}
    o_t=
    \begin{cases}
        \mathcal{R}_{\mathcal{M}}(\phi_t), & z_t=\text{\Retrieve},\\
        \mathcal{C}_{\mathcal{H}}(\phi_t), & z_t=\text{\Curate},
    \end{cases}
\end{equation}
where $\mathcal{R}_{\mathcal{M}}(\cdot)$ denotes semantic-similarity retrieval over the compressed memory bank, while $\mathcal{C}_{\mathcal{H}}(\cdot)$ denotes query-specific runtime curation over the raw history, including evidence retrieval and subsequent processing. In our instantiation, $\mathcal{C}_{\mathcal{H}}(\cdot)$ delegates the retrieved evidence to a heterogeneous pool of external LLMs and VLMs (elaborated in Section~\ref{sec:orchestration}), whereas the orchestrator itself remains text-only and never directly processes raw multimodal inputs (e.g., image pixels). After a variable number of steps, the policy terminates memory curation and produces the final answer $y$.

This process yields a query-adaptive curation trajectory with at most $T_{\max}$ steps:
\begin{equation}
    \tau = (q,a_1,o_1,\ldots,a_T,o_T,y), \quad T \leq T_{\max},
\end{equation}
whose behavior can vary substantially across queries and optimization preferences. We characterize each trajectory by its answer quality $Q(\tau)$ and runtime resource consumption:
\begin{equation}
    \mathbf{R}(\tau)=\big(C(\tau),\,L(\tau)\big),
\end{equation}
where $C(\tau)$ denotes monetary cost and $L(\tau)$ denotes user-facing latency. Given an optimization preference $\boldsymbol{\lambda}$, our goal is to learn an orchestrator policy $\pi_\theta$ that flexibly balances answer quality $Q(\tau)$, monetary cost $C(\tau)$, and latency $L(\tau)$ rather than optimizing task performance alone. 
This formulation casts on-demand memory orchestration as a preference-conditioned sequential decision problem over answer quality and resources, laying the foundation for explicit trade-off optimization.

\subsection{Factorized Memory Curation and Delegation}
\label{sec:orchestration}

\textbf{Factorized curation controls.}
We next instantiate the structured action arguments $\phi_t$ to expose fine-grained control over runtime memory curation. Rather than relying on a fixed processing pipeline or predefined budget tiers, MemPilot factorizes each action into several controllable dimensions, allowing the orchestrator to adapt \emph{what evidence to retrieve}, \emph{how much to retrieve}, \emph{how it should be curated}, \emph{which model should process it}, and \emph{whether visual evidence is needed}. These decisions jointly provide flexible control over runtime computation and its resulting quality--cost--latency trade-offs.

We factorize the action arguments according to the two action types:
\begin{equation}
    \phi_t=
    \begin{cases}
        (r_t,k_t), & z_t=\text{\Retrieve},\\
        (r_t,k_t,i_t,m_t,v_t), & z_t=\text{\Curate},
    \end{cases}
\end{equation}
where $r_t$ is the retrieval query and $k_t$ specifies the evidence count, allowing the orchestrator to adapt retrieval depth across steps rather than relying on a fixed top-$k$. For \Curate{}, $i_t$ further specifies a query-specific curation instruction, $m_t$ selects the processing backend, and $v_t$ determines whether original visual evidence is included. Importantly, we decouple \emph{what to retrieve} ($r_t$, e.g., ``Lena's dress options") from \emph{how to curate it} ($i_t$, e.g., ``extract the two dress types she considered"), avoiding conflating two semantically distinct operations in a single textual field while keeping retrieval evidence-focused and curation task-directed.

\textbf{Heterogeneous model delegation.}
Beyond deciding what evidence to retrieve and how to curate it, MemPilot also adapts which model performs the runtime curation, since heterogeneous models exhibit distinct capabilities, monetary costs, and latencies, making model choice a natural and effective control dimension for runtime trade-offs. 
Inspired by the LLM routing paradigm~\citep{feng2024graphrouter, zhang2025router}, MemPilot therefore delegates runtime curation to a configurable pool of LLMs and VLMs, allowing the orchestrator to adaptively select a model for each curation step under different optimization preferences. 
To support this routing decision, the orchestrator is given anonymized model IDs together with lightweight profiles describing key characteristics such as parameter scale and supported input modalities. The selected model $m_t$ then processes the retrieved evidence according to $i_t$ and returns a textual observation to the orchestrator. For vision-capable routes, $v_t$ further determines whether raw images should be included in the curation request.

Together, these factorized decisions allow each query to induce its own curation trajectory over evidence depth, processing strategy, model capacity, and modality access, providing flexible trade-off control without requiring a rigid processing pipeline or predefined budget tiers.

\subsection{Resource-Aware Reward Modeling}
\label{sec:reward}

Having exposed multiple controls over runtime curation, we train the orchestrator to use them according to different optimization preferences. For a preference vector $\boldsymbol{\lambda}=(\lambda_Q,\lambda_C,\lambda_L)$ (with $\lambda_Q=1$ throughout this work), we optimize the policy with Group Relative Policy Optimization (GRPO)~\citep{shao2024deepseekmath} over answer quality, monetary cost, and latency.

\noindent\textbf{Resource-aware GRPO.}
For each query, the policy $\pi_\theta$ samples a group of $G$ curation trajectories $\{\tau_i\}_{i=1}^{G}$ and evaluates each trajectory with three objective-specific rewards:
\begin{equation}
    r_i^Q = Q(\tau_i), \qquad
    r_i^C = -C(\tau_i), \qquad
    r_i^L = -L(\tau_i),
\end{equation}
where $Q(\tau_i)$ is the task-specific answer reward (e.g., F1-Score). Monetary cost accounts for token-based inference costs incurred by both orchestrator generation and delegated model calls,
\begin{equation}
    C(\tau_i)=C_{\pi}(\tau_i)+C_{\mathrm{ext}}(\tau_i).
\end{equation}
Latency similarly considers both sources of runtime delay. Since direct wall-clock measurements are sensitive to network and serving fluctuations, we use a proxy estimator
$L(\tau_i)=\widehat{L}_{\pi}(\tau_i)+\widehat{L}_{\mathrm{ext}}(\tau_i)$ to ensure training stability, with more details provided in Appendix~\ref{app:reward_modeling}.

\noindent\textbf{Objective-wise advantage decoupling.}
Since directly combining multiple rewards before group normalization can obscure their individual learning signals, inspired by~\citep{liu2026gdpo}, we instead normalize each objective independently within the rollout group:
\begin{equation}
    A_i^j =
    \frac{r_i^j-\mu^j}{\sigma^j+\epsilon_{\mathrm{norm}}},
    \qquad j\in\{Q,C,L\},
\end{equation}
where $\mu^j$ and $\sigma^j$ are the group-wise mean and standard deviation for objective $j$, and $\epsilon_{\mathrm{norm}}$ is a small constant for numerical stability. We then aggregate the normalized advantages according to the preference $\boldsymbol{\lambda}$ and further normalize the combined signal at the batch level:
\begin{equation}
    \bar{A}_i^{\boldsymbol{\lambda}}
    = \lambda_Q A_i^Q + \lambda_C A_i^C + \lambda_L A_i^L,
    \qquad
    A_i^{\boldsymbol{\lambda}}
    = \frac{\bar{A}_i^{\boldsymbol{\lambda}}-\mu_{\bar{A}}}
    {\sigma_{\bar{A}}+\epsilon_{\mathrm{norm}}},
\end{equation}
where $\mu_{\bar{A}}$ and $\sigma_{\bar{A}}$ are the batch-wise mean and standard deviation. The resulting advantage is used in the following clipped GRPO loss:
\begin{equation}
\label{eq:grpo}
    \mathcal{L}_{\mathrm{MemPilot}}
    =
    -\mathbb{E}_{i,k}
    \left[
    \min\!\left(
        \rho_{i,k} A_i^{\boldsymbol{\lambda}},
        \operatorname{clip}(\rho_{i,k},1-\epsilon_{\mathrm{clip}},1+\epsilon_{\mathrm{clip}})
        A_i^{\boldsymbol{\lambda}}
    \right)
    \right]
    + \beta D_{\mathrm{KL}}(\pi_\theta\|\pi_{\mathrm{ref}}),
\end{equation}
where $\rho_{i,k}$ is the probability ratio between the current policy $\pi_\theta$ and the rollout policy $\pi_{\mathrm{old}}$ for token $k$ of trajectory $i$, $\epsilon_{\mathrm{clip}}$ is the clipping coefficient, $\beta$ controls the KL regularization strength, and $\pi_{\mathrm{ref}}$ denotes the reference policy. Varying $\boldsymbol{\lambda}$ yields policies with different quality--cost--latency trade-offs while preserving objective-specific learning signals.

\noindent\textbf{Prefix-based marginal utility.}
Trajectory-level advantages provide the same learning signal across curation stages, obscuring their individual contributions. We therefore probe the rollout after each stage $t$ by truncating subsequent curation and directly generating an answer $\hat{y}_{i,t}$. Let
\begin{equation}
    u_{i,t}=Q(\hat{y}_{i,t}),
    \qquad
    \Delta u_{i,t}=u_{i,t}-u_{i,t-1},
\end{equation}
where $\Delta u_{i,t}$ estimates the marginal answer-quality improvement contributed by stage $t$. Since cost and latency are directly attributable and already captured at the trajectory level, prefix probing focuses only on answer quality to avoid double-counting resource penalties. We use this signal to refine credit assignment for the policy decisions associated with that stage:
\begin{equation}
    \widetilde{A}_{i,k}
    =
    A_i^{\boldsymbol{\lambda}}
    + \eta\,\Delta u_{i,t},
    \qquad
    k\in\mathcal{S}_{i,t},
\end{equation}
where $\mathcal{S}_{i,t}$ contains the policy reasoning and action tokens that precede the corresponding curation response, and $\eta$ controls the strength of the marginal credit. Other tokens retain the trajectory-level advantage $A_i^{\boldsymbol{\lambda}}$. The stage-refined advantage $\widetilde{A}_{i,k}$ replaces $A_i^{\boldsymbol{\lambda}}$ in Eq.~\ref{eq:grpo} for $k\in\mathcal{S}_{i,t}$, incurring little overhead due to short answer completions and largely shared prefix KV-cache reuse.

\section{Experiment}

\subsection{Setups}

\textbf{Benchmarks and Metrics.}
We mainly evaluate MemPilot in multimodal settings involving both textual and visual interaction histories. Specifically, we use five multimodal agent-memory benchmarks: Mem-Gallery~\citep{bei2026mem}, WorldMemArena~\citep{liu2026worldmemarena}, H2HMem~\citep{zhu2026h2hmem}, MemEye~\citep{guo2026memeye}, and MemLens~\citep{ren2026memlens}, with MemEye and MemLens further serving as out-of-distribution evaluation sets.
We evaluate answer quality using F1 score and an LLM-judge score (L-J), with GPT-4o-mini as the judge model. Cost measures the total token-based monetary cost incurred by both the orchestrator and delegated external models. Latency measures their total inference latency in seconds. Appendix~\ref{app:reward_modeling} and~\ref{app:exp_eval} provide further details.

\textbf{Baselines.}
We organize baselines into two groups according to whether they support explicit trade-off control. The first group contains fixed-configuration methods used for the main comparison. It includes \emph{(1) text-centric memory systems} including \textbf{A-MEM}~\citep{xu2025mem} and \textbf{SimpleMem}~\citep{liu2026simplemem}, \emph{(2) training-based memory methods} including \textbf{GAM}~\citep{yan2025general}, \textbf{AgeMem}~\citep{yu2026agentic}, and \textbf{Mem-T}~\citep{yue2026mem}, and \emph{(3) multimodal memory systems} including \textbf{MIRIX}~\citep{wang2025mirix}, \textbf{MemVerse}~\citep{liu2025memverse}, and \textbf{M2A}~\citep{feng2026m2a}. The second group contains \emph{(4) methods that explicitly expose performance--efficiency trade-offs}, including \textbf{LightMem}~\citep{fang2025lightmem}, \textbf{BudgetMem}~\citep{zhang2026learning}, and \textbf{OmniSimpleMem}~\citep{liu2026omni}. Section~\ref{exp:tradeoffs} compares these methods separately through frontier analysis.

\subsection{Implementation Details}
\label{sec:implementation}

MemPilot is implemented based on VeRL\footnote{\url{https://github.com/volcengine/verl}}, with Qwen3-4B-Instruct~\citep{yang2025qwen3} as the default orchestrator policy. To disentangle gains from RL-enhanced answer generation, we additionally report a decoupled-answer evaluation where the curated context is passed to a separate, non-RL-trained answer model, using Qwen3-VL-4B-Instruct and Qwen3.5-9B. External LLMs and VLMs are accessed through OpenRouter\footnote{\url{https://openrouter.ai}}, with the heterogeneous model pool specified in Appendix~\ref{app:model_pool}. We segment raw interaction histories into 256-token chunks and use Qwen3-Embedding-0.6B~\citep{zhang2025qwen3} for semantic retrieval. Unless otherwise specified, the query-agnostic memory bank is constructed with LLMLingua-2~\citep{pan2024llmlingua} at a 0.4 compression ratio for policy training and main evaluation. We further examine compatibility with memory banks produced by other agent memory systems in Section~\ref{exp:discussion}. We retrieve at most five evidence items per step for both MemPilot and applicable baselines. Since memory items from different systems can vary substantially in length, we directly compare monetary cost and latency rather than memory-item counts. Remaining hyperparameters are provided in Appendix~\ref{app:hyperparameters}.

For training, we adopt a 7/1/2 train/validation/test split. We merge the training portions of Mem-Gallery, WorldMemArena, and H2HMem into a joint training set and evaluate in-domain performance on their held-out test sets. We further use MemLens and MemEye exclusively for out-of-domain evaluation to assess generalization to unseen memory distributions. The same data protocol is applied to all applicable baselines, with detailed dataset statistics provided in Appendix~\ref{app:exp_eval}. For the main comparison, we evaluate three representative MemPilot preferences: \textbf{Perf}, \textbf{Bal}, and \textbf{Cost}, prioritizing answer quality, balanced quality--resource trade-offs, and inference cost, respectively. Additional preference settings are explored in Section~\ref{exp:tradeoffs}.

\subsection{Main Results}

\begin{table*}[t]
\centering
\caption{
\textbf{Overall comparison on Mem-Gallery, WorldMemArena, H2HMem, MemEye, and MemLens}.
We report F1 score (\textbf{F1}), LLM-judge score (\textbf{L-J}), and inference cost (\textbf{Cost}), with latency analyzed separately in Section~\ref{exp:tradeoffs}.
Within each column, the best result is boldfaced and highlighted with a darker shade, while the second-best result is highlighted with a lighter shade.
}
\label{tab:main_results}

\setlength{\tabcolsep}{2.6pt}
\renewcommand{\arraystretch}{1.12}
\scriptsize

\resizebox{\textwidth}{!}{
\begin{tabular}{l*{5}{ccc}}
\toprule

\multirow{2}{*}{\textbf{Method}}
&
\multicolumn{3}{c}{\textbf{Mem-Gallery}}
&
\multicolumn{3}{c}{\textbf{WorldMemArena}}
&
\multicolumn{3}{c}{\textbf{H2HMem}}
&
\multicolumn{3}{c}{\textbf{MemEye}$^{\dagger}$}
&
\multicolumn{3}{c}{\textbf{MemLens}$^{\dagger}$}
\\

\cmidrule(lr){2-4}
\cmidrule(lr){5-7}
\cmidrule(lr){8-10}
\cmidrule(lr){11-13}
\cmidrule(lr){14-16}

&
\cellcolor{f1header}\textbf{F1}$\uparrow$
&
\cellcolor{ljheader}\textbf{L-J}$\uparrow$
&
\cellcolor{costheader}\textbf{Cost}$\downarrow$
&
\cellcolor{f1header}\textbf{F1}$\uparrow$
&
\cellcolor{ljheader}\textbf{L-J}$\uparrow$
&
\cellcolor{costheader}\textbf{Cost}$\downarrow$
&
\cellcolor{f1header}\textbf{F1}$\uparrow$
&
\cellcolor{ljheader}\textbf{L-J}$\uparrow$
&
\cellcolor{costheader}\textbf{Cost}$\downarrow$
&
\cellcolor{f1header}\textbf{F1}$\uparrow$
&
\cellcolor{ljheader}\textbf{L-J}$\uparrow$
&
\cellcolor{costheader}\textbf{Cost}$\downarrow$
&
\cellcolor{f1header}\textbf{F1}$\uparrow$
&
\cellcolor{ljheader}\textbf{L-J}$\uparrow$
&
\cellcolor{costheader}\textbf{Cost}$\downarrow$
\\

\midrule

% =========================================================
% Qwen3-4B
% =========================================================

\rowcolor{panelgray}
\multicolumn{16}{l}{\textbf{Answer Model: Qwen3-VL-4B-Instruct}}
\\

A-Mem
& \cellcolor{f1second}54.24 & 51.45 & 2.8e-2
& \cellcolor{f1second}21.38 & 47.73 & 1.8e-1
& 23.27 & 23.88 & 3.6e-2
& 15.20 & 20.28 & 1.3e-2
& 22.99 & 21.15 & 2.6e-2
\\

SimpleMem
& 27.32 & 32.00 & 1.9e-2
& 15.58 & 26.36 & 4.5e-2
& 16.67 & 20.38 & 2.2e-2
& 14.69 & 16.87 & \cellcolor{costbest}\textbf{4.6e-3}
& 18.46 & 20.23 & 2.3e-3
\\

MIRIX
& 36.70 & 49.45 & 3.3e-1
& 19.23 & 54.09 & 7.8e-1
& 19.81 & 25.56 & 3.8e-1
& 12.92 & 21.23 & 2.4e-1
& 15.91 & 13.73 & 8.4e-2
\\

M2A
& 39.05 & 50.73 & 8.3e-1
& 20.14 & 54.49 & 1.1e+0
& \cellcolor{f1second}27.21 & \cellcolor{ljsecond}33.19 & 9.1e-1
& 12.26 & 17.86 & 8.5e-1
& 15.23 & 17.20 & 3.0e-1
\\

MemVerse
& 30.92 & 44.27 & 4.7e-1
& 21.22 & \cellcolor{ljsecond}60.91 & 1.1e+0
& 20.03 & 26.04 & 6.0e-1
& 15.28 & 19.95 & 3.2e-1
& 18.99 & 16.91 & 1.4e-1
\\

GAM
& 43.46 & 46.36 & 1.5e-2
& 13.33 & 35.68 & 3.9e-2
& 23.87 & 23.96 & 2.3e-2
& 12.04 & 19.64 & 2.4e-2
& \cellcolor{f1second}23.43 & 20.81 & 2.2e-3
\\

AgeMem
& 38.38 & 39.36 & 3.0e-2
& 19.45 & 42.95 & 5.4e-2
& 16.46 & 16.11 & 3.6e-2
& 10.48 & 11.86 & 1.9e-2
& 15.22 & 14.19 & 2.5e-3
\\

Mem-T
& 36.18 & 34.55 & 1.9e-2
& 21.14 & 43.27 & 7.0e-2
& 23.15 & 23.85 & 2.3e-2
& \cellcolor{f1best}\textbf{20.89} & \cellcolor{ljsecond}22.78 & 9.0e-3
& 17.22 & 18.01 & 4.6e-3
\\

\midrule

\textbf{MemPilot-Perf}
& 53.73 & \cellcolor{ljbest}\textbf{65.82} & 4.3e-2
& \cellcolor{f1best}\textbf{22.01} & \cellcolor{ljbest}\textbf{63.41} & 5.1e-2
& \cellcolor{f1best}\textbf{30.32} & \cellcolor{ljbest}\textbf{35.97} & 6.0e-2
& \cellcolor{f1second}15.34 & \cellcolor{ljbest}\textbf{25.07} & 5.0e-2
& \cellcolor{f1best}\textbf{30.58} & \cellcolor{ljbest}\textbf{27.17} & 5.5e-4
\\

\textbf{MemPilot-Bal}
& \cellcolor{f1best}\textbf{57.26} & \cellcolor{ljsecond}64.27 & \cellcolor{costsecond}1.4e-2
& 20.71 & 50.00 & \cellcolor{costsecond}1.3e-2
& 24.88 & 28.96 & \cellcolor{costsecond}1.6e-2
& 13.46 & 22.37 & 1.0e-2
& 21.99 & \cellcolor{ljsecond}21.39 & \cellcolor{costsecond}1.8e-4
\\

\textbf{MemPilot-Cost}
& 32.67 & 39.09 & \cellcolor{costbest}\textbf{8.2e-3}
& 6.25 & 15.00 & \cellcolor{costbest}\textbf{7.0e-3}
& 15.18 & 14.58 & \cellcolor{costbest}\textbf{8.5e-3}
& 8.13 & 12.94 & \cellcolor{costsecond}5.6e-3
& 7.63 & 6.36 & \cellcolor{costbest}\textbf{1.2e-4}
\\

\midrule
\midrule

% =========================================================
% Qwen3.5-9B
% =========================================================

\rowcolor{panelgray}
\multicolumn{16}{l}{\textbf{Answer Model: Qwen3.5-9B}}
\\

A-Mem
& 44.68 & 53.82 & 5.5e-2
& 20.33 & 54.09 & 2.6e-1
& 24.78 & 31.46 & 5.9e-2
& 13.04 & 19.21 & 2.2e-2
& 21.46 & 19.58 & 3.6e-2
\\

SimpleMem
& 25.26 & 30.73 & 3.4e-2
& 10.65 & 19.77 & 6.0e-2
& 19.14 & 20.28 & 2.6e-2
& 11.27 & 14.69 & \cellcolor{costbest}\textbf{6.9e-3}
& 13.42 & 11.56 & 3.6e-3
\\

MIRIX
& 37.13 & 50.64 & 3.3e-1
& 20.59 & 57.05 & 1.2e+0
& 22.80 & 31.94 & 6.9e-1
& 11.18 & \cellcolor{ljsecond}21.48 & 2.8e-1
& 20.34 & \cellcolor{ljsecond}22.11 & 1.1e-1
\\

M2A
& 26.14 & 36.00 & 9.5e+0
& 16.26 & 42.50 & 7.6e+0
& 22.89 & 24.44 & 1.0e+1
& 10.41 & 13.95 & 6.1e+0
& 13.40 & 14.23 & 9.3e-1
\\

MemVerse
& 21.08 & 42.45 & 5.0e-1
& 21.01 & \cellcolor{ljsecond}60.23 & 1.3e+0
& 20.91 & 32.85 & 5.2e-1
& 10.89 & 13.01 & 3.5e-1
& 17.89 & 17.63 & 1.8e-1
\\

GAM
& 38.36 & 37.72 & \cellcolor{costsecond}2.1e-2
& 20.51 & 37.50 & 5.4e-2
& 26.92 & 23.68 & 2.8e-2
& 14.13 & 19.56 & 1.5e-2
& 22.51 & 19.08 & 3.7e-3
\\

AgeMem
& 33.48 & 34.55 & 4.4e-2
& 20.17 & 45.00 & 7.2e-2
& 17.41 & 18.33 & 4.9e-2
& 11.21 & 11.99 & 3.0e-2
& 16.35 & 14.45 & 4.1e-3
\\

Mem-T
& 33.00 & 31.18 & \cellcolor{costsecond}2.1e-2
& \cellcolor{f1second}21.72 & 44.09 & 7.9e-2
& 26.01 & 27.92 & \cellcolor{costsecond}2.3e-2
& 12.88 & 19.05 & 1.3e-2
& \cellcolor{f1second}23.20 & 21.59 & 5.5e-3
\\

\midrule

\textbf{MemPilot-Perf}
& \cellcolor{f1second}48.19 & \cellcolor{ljbest}\textbf{67.64} & 5.0e-2
& \cellcolor{f1best}\textbf{23.55} & \cellcolor{ljbest}\textbf{74.09} & 5.7e-2
& \cellcolor{f1best}\textbf{30.62} & \cellcolor{ljbest}\textbf{37.57} & 6.7e-2
& \cellcolor{f1best}\textbf{15.52} & \cellcolor{ljbest}\textbf{21.83} & 5.4e-2
& \cellcolor{f1best}\textbf{28.08} & \cellcolor{ljbest}\textbf{24.28} & 6.5e-4
\\

\textbf{MemPilot-Bal}
& \cellcolor{f1best}\textbf{53.55} & \cellcolor{ljsecond}67.27 & \cellcolor{costsecond}2.1e-2
& 20.98 & 59.32 & \cellcolor{costsecond}1.9e-2
& \cellcolor{f1second}28.58 & \cellcolor{ljsecond}35.69 & \cellcolor{costsecond}2.3e-2
& \cellcolor{f1second}15.26 & 18.87 & 1.5e-2
& 18.51 & 17.92 & \cellcolor{costsecond}2.8e-4
\\

\textbf{MemPilot-Cost}
& 29.74 & 38.55 & \cellcolor{costbest}\textbf{1.5e-2}
& 8.81 & 21.14 & \cellcolor{costbest}\textbf{1.3e-2}
& 14.43 & 16.39 & \cellcolor{costbest}\textbf{1.6e-2}
& 8.77 & 8.09 & \cellcolor{costsecond}1.0e-2
& 8.08 & 6.36 & \cellcolor{costbest}\textbf{2.2e-4}
\\

\bottomrule
\end{tabular}
}

\vspace{3pt}

\begin{minipage}{0.99\textwidth}
\footnotesize
$^{\dagger}$ denotes out-of-distribution (OOD) evaluation datasets.
\end{minipage}

\end{table*}

As shown in Table~\ref{tab:main_results}, MemPilot spans distinct quality--cost regimes across both models and benchmarks. The results reveal several consistent trends across methods and settings. \emph{(1) Text-centric systems achieve low cost largely by avoiding additional multimodal processing and runtime curation, but this often comes with lower answer quality.} In comparison, MemPilot-Bal generally achieves stronger overall quality at moderate cost, while MemPilot-Cost reaches similarly low-cost regimes on most datasets. \emph{(2) Runtime adaptation does not necessarily translate into controllable trade-offs.} Although GAM, AgeMem, and Mem-T adapt memory processing at inference time, they largely expose a fixed operating regime rather than explicitly controlling the quality--cost balance. In contrast, MemPilot spans multiple operating points. MemPilot-Perf achieves the best L-J score across all five benchmarks under both models, while MemPilot-Bal substantially lowers cost and provides a distinct intermediate quality--cost trade-off. \emph{(3) Multimodal memory processing can be costly without adaptive allocation.} MIRIX, M2A, and MemVerse often incur substantially higher inference cost without consistently translating that expense into better answer quality. In contrast, MemPilot selectively invokes external curation and achieves stronger performance at much lower cost in most settings, with the same trend extending to MemEye and MemLens. \emph{Overall, MemPilot provides controllable quality--cost trade-offs rather than a single performance-oriented operating point.} MemPilot-Perf prioritizes answer quality, MemPilot-Bal offers a balanced operating point, and MemPilot-Cost further reduces resource usage with greater quality sacrifice. We analyze latency and finer-grained performance--cost--latency frontiers in Section~\ref{exp:tradeoffs}.

\begin{figure*}[t]
	\centering
	\includegraphics[width=1.0\linewidth]{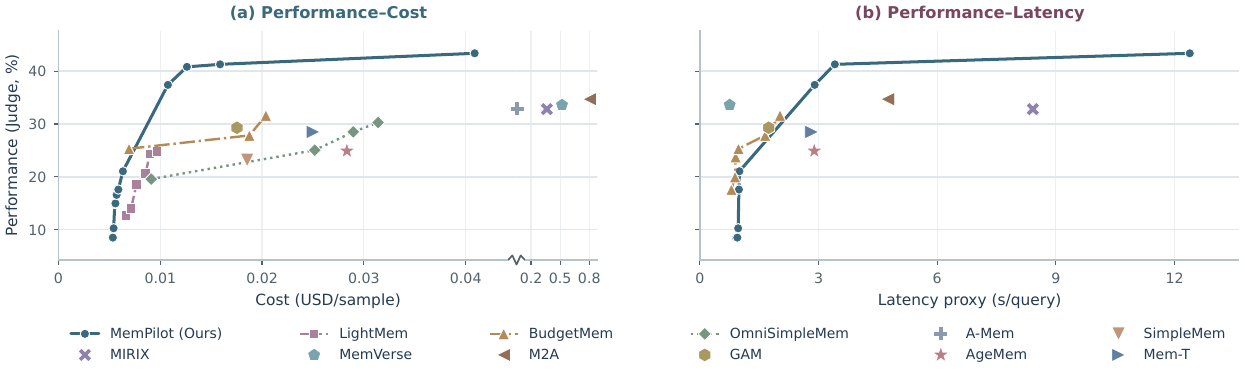}
\caption{\textbf{Performance--cost--latency frontiers.}
Results are macro-averaged over five benchmarks. Lines connect Pareto-optimal points, while faint markers show other evaluated settings. Operating points are obtained by varying optimization preferences or budget settings.}
\label{fig:frontier}
\end{figure*}

\subsection{Exploring Performance--Cost--Latency Frontiers}
\label{exp:tradeoffs}

We further examine fine-grained trade-off frontiers by sweeping MemPilot's optimization preferences. In addition to the main baselines, we include LightMem~\citep{fang2025lightmem}, BudgetMem~\citep{zhang2026learning}, and OmniSimpleMem~\citep{liu2026omni}, which explicitly support performance--efficiency trade-offs. For latency, we report representative fixed-inference baselines together with runtime-adaptive methods, since most query-agnostic systems exhibit nearly unchanged inference behavior and collapse to similar single operating points.

\textbf{Performance--cost.}
As shown in Figure~\ref{fig:frontier}(a), MemPilot forms a stronger frontier across a broad cost range and largely envelopes the competing methods. Compared with trade-off-aware baselines such as LightMem, BudgetMem, and OmniSimpleMem, MemPilot reaches higher quality as additional computation is allocated, while also covering competitive low-cost operating points. Most conventional memory systems expose only a fixed quality--cost point, highlighting the additional flexibility provided by learned runtime orchestration. Multimodal systems such as MIRIX, M2A, and MemVerse generally incur much higher cost due to visual processing, whereas MemPilot reduces unnecessary multimodal computation through on-demand curation and selective model delegation.

\textbf{Performance--latency.}
Figure~\ref{fig:frontier}(b) shows a similarly broad quality--latency frontier. MemPilot spans lightweight to performance-oriented configurations, whereas most baselines expose only a narrow region or a single operating point. Even among runtime-adaptive methods, the achievable latency range remains limited. MemVerse achieves competitive quality at relatively low latency, but lacks explicit trade-off control and incurs substantially higher monetary cost in Figure~\ref{fig:frontier}(a). This further highlights that cost and latency capture complementary resource constraints.

Overall, MemPilot provides a broader, more controllable performance--cost--latency frontier through adaptive runtime orchestration, enabling flexible computation allocation under resource preferences.

\subsection{How Does MemPilot Allocate Runtime Computation?}
\label{exp:behavior}

\begin{figure*}[t]
	\centering
	\includegraphics[width=1.0\linewidth]{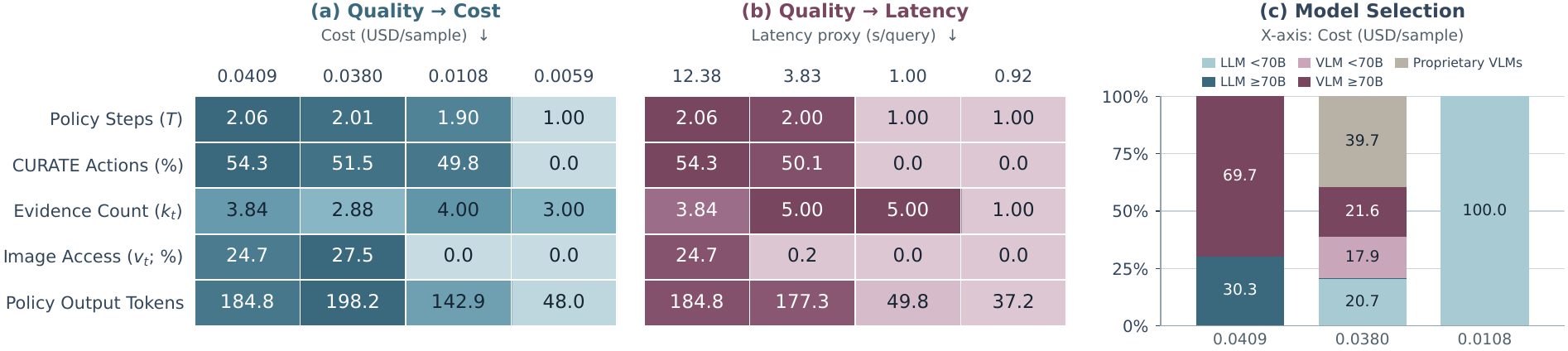}
        \caption{\textbf{Allocation of runtime computation under different optimization preferences.}
        Panels (a,b) start with the same quality-oriented policy, with cost and latency proxy decreasing from left to right, respectively. Panel (c) shows model-selection shares across model groups. Values are averaged across five benchmarks; darker cells indicate larger values within each row.}
\label{fig:policy_controls}
\end{figure*}

To better understand how MemPilot realizes different trade-offs, we analyze its learned runtime curation behavior in Figure~\ref{fig:policy_controls}. Panels (a) and (b) show that as the optimization preference shifts toward lower cost or lower latency, the policy adjusts multiple control dimensions in a coordinated manner rather than relying on a single knob. Moving from left to right, MemPilot generally reduces the number of policy steps, the frequency of \Curate{} actions, image access, and policy output tokens. At the same time, evidence count also changes adaptively, indicating that the policy does not simply retrieve less information, but learns to reconfigure how memory is accessed and processed under different resource preferences.

Panel (c) further shows that model selection is a key mechanism behind this adaptation. Under higher-cost settings, MemPilot relies more heavily on stronger VLMs and LLMs, while lower-cost settings shift computation toward cheaper model groups. Together, these results show that MemPilot learns a coherent allocation strategy over curation depth, modality access, and routed model capacity, which explains its ability to realize controllable performance--cost--latency trade-offs.

\subsection{Discussion}
\label{exp:discussion}

\begin{wraptable}{r}{0.40\textwidth}
    \vspace{-12pt}
    \centering
    \small
    \setlength{\tabcolsep}{5pt}

    \caption{\textbf{Ablation study.}
    Results are averaged across five benchmarks.}
    \label{tab:ablation}

    \begin{tabular}{@{}lcc@{}}
        \toprule
        Variant
        & L-J $\uparrow$
        & Cost $\downarrow$ \\
        \midrule
        \textbf{MemPilot-Bal}
        & \textbf{37.40}
        & 1.1e-2 \\
        \midrule
        w/o LLM/VLM Delegation
        & 14.58
        & 5.6e-3 \\
        w/o Marginal Utility
        & 14.82
        & 5.5e-3 \\
        w/o GDPO
        & 21.94
        & 5.9e-3 \\
        \shortstack[l]{w/o Retrieval--Instruction\\Decoupling}
        & 15.35
        & 5.4e-3 \\
        \bottomrule
    \end{tabular}
    \vspace{-9pt}
\end{wraptable}

\textbf{Ablation Study of Core Design Choices.}
We ablate four key components of MemPilot to examine their contributions to runtime orchestration and policy optimization, with Qwen3-VL-4B-Instruct as an answer model. As shown in Table~\ref{tab:ablation}, removing LLM/VLM delegation reduces the average Judge score from 37.40 to 14.58, highlighting the importance of adaptive external curation. Removing marginal utility or retrieval--instruction decoupling causes similarly large degradation, demonstrating the value of fine-grained credit assignment and separating evidence retrieval from curation instructions. Objective-wise advantage decoupling is also important for multi-objective training, with its removal reducing the Judge score to 21.94. Although all ablated variants consume less inference cost, their substantial quality losses indicate that simply reducing runtime computation does not yield a desirable trade-off.

\begin{wrapfigure}{r}{0.4\textwidth}
    % \vspace{-8pt}
    \centering
    \includegraphics[width=\linewidth]{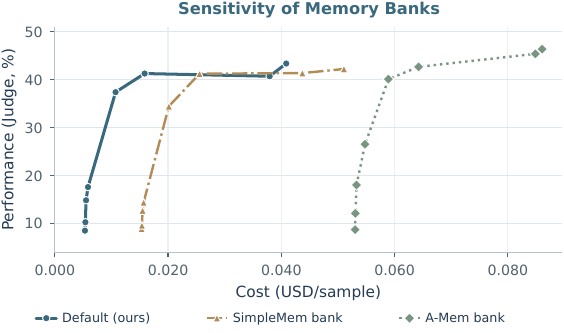}
    \caption{\textbf{Memory-bank compatibility.}
Performance--cost frontiers with different memory-bank initializations.}
    \label{fig:sen_mb}
    \vspace{-7pt}
\end{wrapfigure}

\textbf{Compatibility with Off-the-Shelf Memory Banks.}
To examine whether MemPilot depends on a particular query-agnostic memory construction, we replace its default memory bank with off-the-shelf agent memory banks and evaluate the performance--cost frontiers without retraining using Qwen3-VL-4B-Instruct as an answer model. Figure~\ref{fig:sen_mb} reports averages across five benchmarks. The resulting frontiers are clearly separated: the default bank occupies the lower-cost regime, SimpleMem provides an intermediate profile, while A-MEM incurs higher cost but reaches stronger performance. Each memory bank thus induces a distinct operating regime, while MemPilot improves performance as additional runtime computation is allocated. This demonstrates compatibility with diverse upstream memory systems while preserving their distinct performance--cost characteristics through runtime orchestration.

\section{Conclusion}

We presented \textbf{MemPilot}, a framework for orchestrating on-demand memory curation under controllable performance--cost--latency preferences. MemPilot formulates runtime memory adaptation as a multi-step decision process, combines query-agnostic memory with raw interaction history, and adaptively delegates curation to heterogeneous LLMs and VLMs. We further introduced resource-aware policy optimization with objective-wise advantage decoupling and prefix-based marginal utility for fine-grained credit assignment. Experiments across multimodal agent-memory benchmarks demonstrate that MemPilot achieves favorable trade-off frontiers while adapting computation to different optimization preferences. We hope this work encourages agent memory systems to move beyond performance-only optimization toward resource-aware runtime adaptation.

\bibliography{references}

\clearpage
\appendix
\startcontents[appendix]
\section*{Appendix Contents}
\printcontents[appendix]{}{1}{\setcounter{tocdepth}{2}}
\clearpage
\section*{Appendix}
\section{More Implementation Details}
\label{app:details}

\subsection{Detailed Reward Modeling}
\label{app:reward_modeling}

\noindent\textbf{Cost Reward.}
We account for the inference cost of both the locally hosted orchestrator and the external models. Let $\mathcal{S}_{\pi}(\tau)$ denote the set of orchestrator-generation segments in trajectory $\tau$, and let $\mathcal{E}(\tau)$ denote its external-model calls. The monetary cost in USD is
\begin{equation}
\begin{split}
C_{\pi}(\tau)
&=
10^{-6}
\sum_{s\in\mathcal{S}_{\pi}(\tau)}
\left(
p_{\pi}^{\mathrm{in}}n_{\pi,s}^{\mathrm{in}}
+
p_{\pi}^{\mathrm{out}}n_{\pi,s}^{\mathrm{out}}
\right),\\
C_{\mathrm{ext}}(\tau)
&=
10^{-6}
\sum_{c\in\mathcal{E}(\tau)}
\left(
p_{m_c}^{\mathrm{in}}n_c^{\mathrm{in}}
+
p_{m_c}^{\mathrm{out}}n_c^{\mathrm{out}}
\right),\\
C(\tau)
&=
C_{\pi}(\tau)+C_{\mathrm{ext}}(\tau),
\end{split}
\end{equation}
where $m_c$ is the external model selected for call $c$, and all prices are specified in USD per million tokens. We assign the orchestrator proxy prices
$p_{\pi}^{\mathrm{in}}=0.05$ and
$p_{\pi}^{\mathrm{out}}=0.25$ for \texttt{Qwen3-4B-Instruct}. The external-model prices
$p_m^{\mathrm{in}}$ and $p_m^{\mathrm{out}}$ are listed in
Table~\ref{tab:model_pool}. When the API returns an exact monetary cost, we use it in place of the static token-based estimate.

For orchestrator accounting, the initial prompt is charged once. Each policy-generated token is counted once as output, while newly appended environment tokens, including observation responses, are counted as incremental inputs to the next step generation. Previously processed prompt and assistant-history tokens are not charged again, corresponding to ideal within-trajectory KV-prefix reuse. A direct \Retrieve{} call incurs no external-model cost, although the orchestrator generations before and after that call remain included.

\noindent\textbf{Latency Reward.}
We use a deterministic latency proxy rather than the observed wall-clock latency of individual API requests, which can vary substantially with network conditions and transient serving load. LLM inference latency is roughly influenced by fixed per-request overhead, input-dependent prefill computation, and autoregressive output decoding. Rather than modeling the full hardware- and serving-dependent latency profile, we adopt a deliberately simplified affine proxy that summarizes these effects with a fixed overhead and average per-token contributions for input and output over the operating range considered in our experiments. Accordingly, for an orchestrator-generation segment $s$, we estimate its latency as
\begin{equation}
\widehat{L}_{\pi,s}
=
b_{\pi}
+
a_{\pi}^{\mathrm{in}} n_{\pi,s}^{\mathrm{in}}
+
a_{\pi}^{\mathrm{out}} n_{\pi,s}^{\mathrm{out}},
\end{equation}
where $b_{\pi}$ captures the fixed invocation and scheduling overhead, while
$a_{\pi}^{\mathrm{in}}$ and $a_{\pi}^{\mathrm{out}}$ model the marginal latency of input prefill and autoregressive decoding, respectively. The orchestrator coefficients are
\begin{equation}
\left(
b_{\pi},
a_{\pi}^{\mathrm{in}},
a_{\pi}^{\mathrm{out}}
\right)
=
\left(
0.332,\,
3.0{\times}10^{-6},\,
6.64{\times}10^{-3}
\right),
\end{equation}
measured in seconds, seconds per input token, and seconds per output token, respectively. We apply the same incremental-token accounting used for monetary cost: the initial prompt is counted once, each newly generated orchestrator segment contributes output tokens, and only newly inserted environment content contributes additional input tokens.

For an external-model call $c$ routed to model $m_c$, we estimate
\begin{equation}
\widehat{L}_{c}
=
b_{m_c}
+
a_{m_c}^{\mathrm{in}}n_c^{\mathrm{in}}
+
a_{m_c}^{\mathrm{out}}n_c^{\mathrm{out}}
+
a_{m_c}^{\mathrm{img}}n_c^{\mathrm{img}},
\end{equation}
where the image term is zero for text-only LLMs. Table~\ref{tab:model_pool} reports the model-specific coefficients.

Calls emitted in the same assistant turn are executed in parallel and form one latency stage. We therefore take the maximum latency within each parallel stage and sum across sequential stages:
\begin{equation}
\widehat{L}_{\mathrm{ext}}(\tau)
=
\sum_{t=1}^{T}
\max_{c\in\mathcal{E}_t(\tau)}
\widehat{L}_{c},
\end{equation}
where $\mathcal{E}_t(\tau)$ contains the external calls made in stage $t$. The total trajectory latency is
\begin{equation}
\widehat{L}(\tau)
=
\sum_{s\in\mathcal{S}_{\pi}(\tau)}
\widehat{L}_{\pi,s}
+
\widehat{L}_{\mathrm{ext}}(\tau).
\end{equation}

We calibrate the external-model profiles using repeated non-streaming OpenRouter requests. Each model is evaluated through a fixed backend route with provider fallback disabled, and cache-busting request content is used to avoid identical-request cache hits. Controlled short-input, long-input, long-output, and image-input cases isolate the fixed, per-input-token, per-output-token, and per-image terms. We use median measurements to reduce sensitivity to transient outliers and constrain all coefficients to be non-negative.

\noindent\textbf{Overall.}
Let $Q(\tau)\in[0,1]$ denote the task-specific answer-quality score and let
$g(\tau)\in\{0,1\}$ indicate whether the trajectory satisfies the complete
output-format contract. We apply the format gate to obtain the task reward:
\begin{equation}
r^{\mathrm{task}}(\tau)
=
\begin{cases}
Q(\tau), & g(\tau)=1,\\
-1, & g(\tau)=0.
\end{cases}
\end{equation}
Thus, a format-valid trajectory receives its answer-quality score, whereas a
format-invalid trajectory receives a fixed reward of $-1$.

The format gate checks that the trajectory contains exactly one non-empty final
\texttt{<answer>...</answer>} block with no trailing output, that every action has balanced tags and parseable JSON arguments, and that all required action fields are present. Action-side validation failures, such as invalid model or memory identifiers and overlength arguments, also invalidate the trajectory. Since memory evidence is required for the evaluated tasks, producing a final answer without invoking either \Retrieve{} or \Curate{} is also treated as invalid.

Format-invalid and padding trajectories are excluded from both cost and latency comparison. Their resource rewards and resource advantages are set to zero, preventing a malformed or no-tool trajectory from obtaining a favorable resource signal merely because it incurs little cost or latency. For a format-valid trajectory, however, resource rewards remain active even when $Q(\tau)=0$; cost and latency are therefore not multiplied or gated by answer performance. This preserves the independence of the three optimization objectives.

The resulting task, cost, and latency components are normalized and combined at the advantage level as described in Section~\ref{sec:reward}. Prefix-based marginal-utility probes use the ungated answer-quality score because their synthetic outputs intentionally omit the main trajectory's action structure. Probe generations are used only for credit assignment and are excluded from monetary cost, latency, and external-call statistics.

\subsection{Heterogeneous Model Pool}
\label{app:model_pool}

We list the heterogeneous model pool in Table~\ref{tab:model_pool}. All models are accessed through OpenRouter\footnote{\url{https://openrouter.ai/}}. For each model $m$, $p_m^{\mathrm{in}}$ and $p_m^{\mathrm{out}}$ denote the reference input and output prices in USD per million tokens. The latency profile is represented as
$\boldsymbol{\phi}_m=(b_m,a_m^{\mathrm{in}},a_m^{\mathrm{out}})$
for text-only LLMs and
$\boldsymbol{\phi}_m=(b_m,a_m^{\mathrm{in}},a_m^{\mathrm{out}},a_m^{\mathrm{img}})$
for VLMs. Here, $b_m$ is the base latency in seconds per call, while
$a_m^{\mathrm{in}}$, $a_m^{\mathrm{out}}$, and $a_m^{\mathrm{img}}$
are measured in seconds per input token, output token, and input image, respectively.
These coefficients are fixed latency proxies rather than unstable live wall-clock measurements.
Their use and calibration are detailed in Section~\ref{app:reward_modeling}.

\begin{table*}[h]
\centering
\caption{
\textbf{Heterogeneous model pool used in our experiments}.
Input and output prices, denoted by $p_m^{\mathrm{in}}$ and
$p_m^{\mathrm{out}}$, are in USD per million tokens.
The latency profile is ordered as
$(b_m,a_m^{\mathrm{in}},a_m^{\mathrm{out}})$ for LLMs and
$(b_m,a_m^{\mathrm{in}},a_m^{\mathrm{out}},a_m^{\mathrm{img}})$ for VLMs.
}
\label{tab:model_pool}
\small
\renewcommand{\arraystretch}{1.08}
\resizebox{\linewidth}{!}{%
\begin{tabular}{@{}lccc@{}}
\toprule
Model & $p_m^{\mathrm{in}}$ & $p_m^{\mathrm{out}}$
& Latency profile $\boldsymbol{\phi}_m$ \\
\midrule
\multicolumn{4}{@{}l}{\textit{Text-only LLMs}} \\

\texttt{Llama-3.2-3B-Instruct}
& 0.05 & 0.33
& $(1.231,\;4.8{\times}10^{-5},\;5.54{\times}10^{-3})$ \\

\texttt{Llama-3.1-8B-Instruct}
& 0.05 & 0.08
& $(0.439,\;3.6{\times}10^{-5},\;4.99{\times}10^{-3})$ \\

\texttt{Llama-3.3-70B-Instruct}
& 0.13 & 0.40
& $(0.475,\;9.2{\times}10^{-5},\;2.194{\times}10^{-2})$ \\

\texttt{Qwen3-30B-A3B-Instruct-2507}
& 0.10 & 0.30
& $(0.524,\;4.5{\times}10^{-5},\;5.33{\times}10^{-3})$ \\

\texttt{Qwen3-Next-80B-A3B-Instruct}
& 0.10 & 1.10
& $(0.517,\;5.2{\times}10^{-5},\;1.126{\times}10^{-2})$ \\

\texttt{Qwen3-235B-A22B-2507}
& 0.09 & 0.55
& $(0.797,\;5.0{\times}10^{-5},\;5.88{\times}10^{-2})$ \\

\addlinespace
\multicolumn{4}{@{}l}{\textit{Vision-language models}} \\

\texttt{Qwen3-VL-8B-Instruct}
& 0.117 & 0.455
& $(0.332,\;3.0{\times}10^{-6},\;6.64{\times}10^{-3},\;0.225)$ \\

\texttt{Qwen3-VL-30B-A3B-Instruct}
& 0.15 & 0.60
& $(0.438,\;6.6{\times}10^{-5},\;5.327{\times}10^{-2},\;0.052)$ \\

\texttt{Qwen3-VL-235B-A22B-Instruct}
& 0.21 & 1.90
& $(0.588,\;8.5{\times}10^{-5},\;5.293{\times}10^{-2},\;1.722)$ \\

\texttt{Gemini-2.5-Flash-Lite}
& 0.10 & 0.40
& $(0.920,\;6.7{\times}10^{-5},\;3.97{\times}10^{-3},\;1.097)$ \\

\texttt{Claude-Haiku-4.5}
& 1.00 & 5.00
& $(1.189,\;3.5{\times}10^{-5},\;1.151{\times}10^{-2},\;0.904)$ \\

\texttt{GPT-5.6-Luna}
& 0.50 & 3.00
& $(1.697,\;8.8{\times}10^{-5},\;8.09{\times}10^{-3},\;0.282)$ \\
\bottomrule
\end{tabular}%
}
\end{table*}

\subsection{Prefix-Based Marginal-Utility Credit Assignment}
\label{app:marginal_utility}

Algorithm~\ref{alg:marginal_utility} provides the implementation details of the prefix-based credit assignment introduced in Section~\ref{sec:reward}. For a trajectory $\tau_i$ containing $K_i$ causal curation stages, we record the pre-curation prefix $C_{i,0}$ and the prefix $C_{i,t}$ after the evidence returned by stage $t$ has been incorporated. Multiple curation operations initiated concurrently at the same decision point are treated as one stage because their returned evidence is observed jointly.

Using the same rollout policy, we greedily generate a short answer from each prefix without performing any subsequent memory curation. These $K_i+1$ probes yield answer-quality scores $u_{i,0},\ldots,u_{i,K_i}$, from which the signed marginal utility of each stage is computed as
\begin{equation}
\Delta u_{i,t}=u_{i,t}-u_{i,t-1}.
\end{equation}
The resulting credit is added only to the policy reasoning and action tokens associated with that stage. The probes are auxiliary training computations and are excluded from monetary-cost, latency, and delegated-model invocation accounting. They are submitted to the same rollout engine, allowing automatic prefix caching to reuse shared prefixes when available.

\begin{algorithm}[h]
\caption{Prefix-based marginal-utility credit assignment.}
\label{alg:marginal_utility}
\begin{algorithmic}[1]
\Require Completed rollout $\tau_i$ with $K_i$ causal curation stages
\Require Prefixes $\{C_{i,t}\}_{t=0}^{K_i}$ and policy-token spans
         $\{\mathcal{S}_{i,t}\}_{t=1}^{K_i}$
\Require Ground-truth answer $y_i$, trajectory-level advantage
         $A_i^{\boldsymbol{\lambda}}$, marginal weight $\eta$,
         and probe-length limit $T_{\max}$
\Ensure Stage-refined token advantages
        $\{\widetilde{A}_{i,k}\}$

\State $\widetilde{A}_{i,k}
       \gets A_i^{\boldsymbol{\lambda}}$
       for every policy token $k$ in $\tau_i$

\If{$K_i=0$ \textbf{or} $\tau_i$ is format-invalid}
    \State \Return $\{\widetilde{A}_{i,k}\}$
\EndIf

\For{$t=0,\ldots,K_i$ \textbf{in parallel}}
    \State $\hat{y}_{i,t}
    \gets
    \Call{GreedyAnswer}
    {\pi_{\mathrm{old}},
     C_{i,t}\mathbin{\|}\texttt{<answer>},
     T_{\max}}$
    \Comment{No subsequent memory curation}
    \State $u_{i,t}\gets Q(\hat{y}_{i,t},y_i)$
\EndFor

\If{any answer probe fails}
    \State \Return $\{\widetilde{A}_{i,k}\}$
\EndIf

\For{$t=1,\ldots,K_i$}
    \State $\Delta u_{i,t}
    \gets u_{i,t}-u_{i,t-1}$
    \ForAll{$k\in\mathcal{S}_{i,t}$}
        \State $\widetilde{A}_{i,k}
        \gets
        \widetilde{A}_{i,k}
        +\eta\,\Delta u_{i,t}$
    \EndFor
\EndFor

\State \Return $\{\widetilde{A}_{i,k}\}$
\end{algorithmic}
\end{algorithm}

Here, $C_{i,0}$ provides the pre-curation baseline required to estimate the contribution of the first curation stage.

\subsection{Evaluation Details}
\label{app:exp_eval}

\noindent\textbf{Memory preprocessing and indexing.}
We serialize each multimodal conversation in chronological order and partition it into approximately 256-token text chunks. Dialogue turns are treated as atomic units: a turn is never split across chunks, and its attached images, image identifiers, and captions remain associated with the same chunk. An oversized individual turn is retained as a single chunk. For WorldMemArena~\citep{liu2026worldmemarena}, the memory available to each question is additionally restricted to the cumulative conversation prefix visible at its corresponding checkpoint.

We construct separate dense indexes for the raw and compressed memory corpora. Query-agnostic compression is performed independently for chunk text and image captions using LLMLingua-2~\citep{pan2024llmlingua} with a default compression rate of 0.4, while memory and image identifiers remain unchanged. Document embeddings are computed offline using Qwen3-Embedding-0.6B~\citep{zhang2025qwen3}, normalized, stored in float16 format, and reused throughout evaluation. At runtime, only the policy-generated retrieval query is encoded, after which evidence is ranked by embedding similarity. The compressed-memory and raw-memory indexes are searched independently, with at most five retrieved items per curation operation. Image captions participate in textual retrieval, whereas image pixels are made available only when an image-capable external model is selected. Reference answers are never used during preprocessing, indexing, or retrieval.

\noindent\textbf{Benchmarks and data splits.}
Our in-distribution evaluation covers Mem-Gallery~\citep{bei2026mem}, the lifelong subset of WorldMemArena~\citep{liu2026worldmemarena}, and both the dyadic and multiparty variants of H2HMem~\citep{zhu2026h2hmem}. Each dataset is split at the conversation level into 70\% training, 10\% validation, and 20\% test data, ensuring that questions derived from the same conversation never appear in different splits. WorldMemArena is additionally stratified by category and scenario. The source datasets are split independently before their training and validation portions are merged, and their source identities are preserved for dataset-wise reporting.

For out-of-distribution evaluation, we use the complete open-answer portion of MemEye~\citep{guo2026memeye}, containing 371 questions across eight scenarios, and the official 32K memory-agent subset of MEMLENS~\citep{ren2026memlens}. MemEye's multiple-choice variant is excluded. MEMLENS answer-refusal questions are removed, leaving 173 of the 195 agent-subset questions. Answer-refusal questions are likewise excluded from Mem-Gallery and H2HMem. For H2HMem, we additionally remove dialogues without usable memory content and Conflict Detection instances whose reference labels cannot satisfy the benchmark's required standalone Yes/No answer format. Neither OOD benchmark contributes data to policy training or model selection.

\begin{table*}[h]
\centering
\caption{
\textbf{Benchmark statistics after applying our data-integrity and answer-refusal filters}.
An instance denotes an independently partitioned memory history; for MEMLENS,
each question-specific 32K history constitutes one instance.
IID datasets are split at the instance level, whereas OOD benchmarks are used
exclusively for testing.
}
\label{tab:benchmark_statistics}
\small
\setlength{\tabcolsep}{5pt}
\renewcommand{\arraystretch}{1.08}
\begin{tabular}{@{}llrrrrrr@{}}
\toprule
& &
\multicolumn{2}{c}{Train} &
\multicolumn{2}{c}{Validation} &
\multicolumn{2}{c}{Test} \\
\cmidrule(lr){3-4}
\cmidrule(lr){5-6}
\cmidrule(l){7-8}
Benchmark & Setting &
\#Inst. & \#QA &
\#Inst. & \#QA &
\#Inst. & \#QA \\
\midrule
Mem-Gallery
& IID
& 14 & 1,062
& 2 & 190
& 4 & 275 \\

WorldMemArena-Lifelong
& IID
& 27 & 1,485
& 3 & 165
& 8 & 440 \\

H2HMem-Dyadic
& IID
& 13 & 1,122
& 2 & 174
& 4 & 327 \\

H2HMem-Multiparty
& IID
& 3 & 97
& 1 & 33
& 1 & 33 \\

\cmidrule(lr){1-8}
\textit{IID total}
&
& \textbf{57} & \textbf{3,766}
& \textbf{8} & \textbf{562}
& \textbf{17} & \textbf{1,075} \\

\midrule
MemEye-Open
& OOD
& -- & --
& -- & --
& 8 & 371 \\

MEMLENS-32K-Agent
& OOD
& -- & --
& -- & --
& 173 & 173 \\

\cmidrule(lr){1-8}
\textit{OOD total}
&
& -- & --
& -- & --
& \textbf{181} & \textbf{544} \\

\bottomrule
\end{tabular}
\end{table*}

\noindent\textbf{Inference.}
We evaluate one deterministic trajectory per question using greedy decoding with temperature 0. The maximum prompt and response lengths are both 4,096 tokens. Each trajectory permits at most six sequential curation rounds and at most two concurrent independent curation operations. All policy-facing memory-access settings are kept identical to those used during the corresponding training run.

\noindent\textbf{LLM-as-a-judge evaluation.}
We additionally evaluate semantic answer correctness using
\texttt{openai/gpt-4o-mini}. We preserve each benchmark's official judge role, correctness rubric, score semantics, and output convention. The only sample-dependent inputs presented to the judge are the predicted answer and the ground-truth reference answer.

Minor answer-only adaptations are applied when an official prompt requires information unavailable to common baseline outputs. Specifically, we remove the question field from all judge inputs; remove Key Memory Points from WorldMemArena because they evaluate evidence coverage rather than final-answer correctness; convert H2HMem's session-batched evaluation into itemwise evaluation; and remove the previous answer from MEMLENS Knowledge Update prompts because correctness is determined by the current reference answer. Mem-Gallery and H2HMem use five-level scores in
$\{0,0.25,0.5,0.75,1\}$. WorldMemArena classifies each answer as
\textsc{Correct}, \textsc{Hallucination}, or \textsc{Omission}, with only
\textsc{Correct} receiving score 1. MemEye follows its official score quantization to
$\{0,0.5,1\}$, and MEMLENS uses type-specific binary rubrics. Judge-model token usage is retained for audit but excluded from the system's cost, latency, and invocation metrics.

\noindent\textbf{Policy--answer model decoupling.}
For controlled comparisons that separate memory orchestration from answer-generation ability, we optionally remove the trained policy's final answer while retaining the query and the memory evidence collected along its curation trajectory. This information is provided to a separately specified, untrained answer model using a benchmark-specific QA prompt. Policy reasoning, routing arguments, system instructions, and reference answers are excluded from this prompt. The replacement model's cost and latency are added to the corresponding resource totals, but its common final-answer invocation is not counted as a routed external-model invocation.

\noindent\textbf{Aggregation and resource reporting.}
F1, EM, LLM-judge score, and QA latency are macro-averaged over questions. Let $\mathcal{Q}$ denote the evaluated questions and let $N_{\mathrm{sample}}$ be the number of unique dataset--conversation pairs. We aggregate resource metrics as
\begin{equation}
\mathrm{Cost}_{\mathrm{QA}}
=
\frac{\sum_{q\in\mathcal{Q}} C(\tau_q)}
     {N_{\mathrm{sample}}},
\qquad
\mathrm{Latency}_{\mathrm{QA}}
=
\frac{\sum_{q\in\mathcal{Q}} \widehat{L}(\tau_q)}
     {|\mathcal{Q}|},
\qquad
\#\mathrm{Calls}_{\mathrm{QA}}
=
\frac{\sum_{q\in\mathcal{Q}} N_{\mathrm{ext}}(\tau_q)}
     {N_{\mathrm{sample}}},
\end{equation}
where $N_{\mathrm{ext}}(\tau_q)$ counts only attempted routed external-LLM invocations. It excludes local policy generation, retrieval-only operations, judge-model requests, optional common answer-model generation, and offline memory construction. QA cost includes policy inference, routed external models, and the optional decoupled answer model. Section~\ref{app:reward_modeling} provides detailed cost and latency definitions.

\subsection{Hyperparameters}
\label{app:hyperparameters}

We summarize the hyperparameters used in our main experiments in Table~\ref{tab:hyperparameters}.

\begin{table}[h]
\centering
\caption{\textbf{Hyperparameters used in the main experiments}.}
\label{tab:hyperparameters}
\small
\renewcommand{\arraystretch}{1.08}
\begin{tabular}{@{}ll@{}}
\toprule
Hyperparameter & Value \\
\midrule
Policy model & \texttt{Qwen3-4B-Instruct-2507} \\
Embedding model & \texttt{Qwen3-Embedding-0.6B} \\
Maximum retrieved items per call & 5 \\
Rollouts per query $G$ & 4 \\
Rollout temperature & 1.0 \\
Training batch size & 32 \\
Learning rate & $1\times10^{-6}$ \\
Training epochs & 3 \\
Maximum prompt length & 4,096 tokens \\
Maximum response length & 4,096 tokens \\
Maximum context length & 8,192 tokens \\
Maximum user/assistant turns & 6 / 6 \\
Maximum parallel tool calls & 2 \\
Maximum external-model output & 512 tokens \\
External-model temperature & 0.0 \\
GRPO clipping coefficient $\epsilon_{\mathrm{clip}}$ & 0.2 \\
KL-loss coefficient $\beta$ & 0.001 \\
Marginal-utility weight $\eta$ & 0.5 \\
\bottomrule
\end{tabular}
\end{table}

\section{More Experimental Results}
\label{app:results}

\begin{figure*}[t]
    \centering

    \includegraphics[height=0.155\textheight,keepaspectratio]{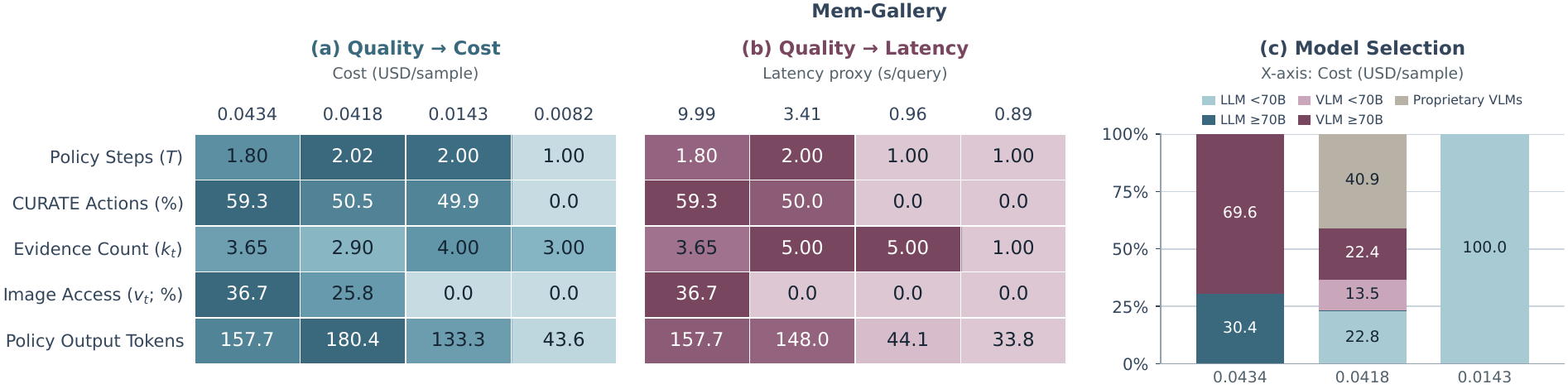}

    \vspace{2pt}

    \includegraphics[height=0.155\textheight,keepaspectratio]{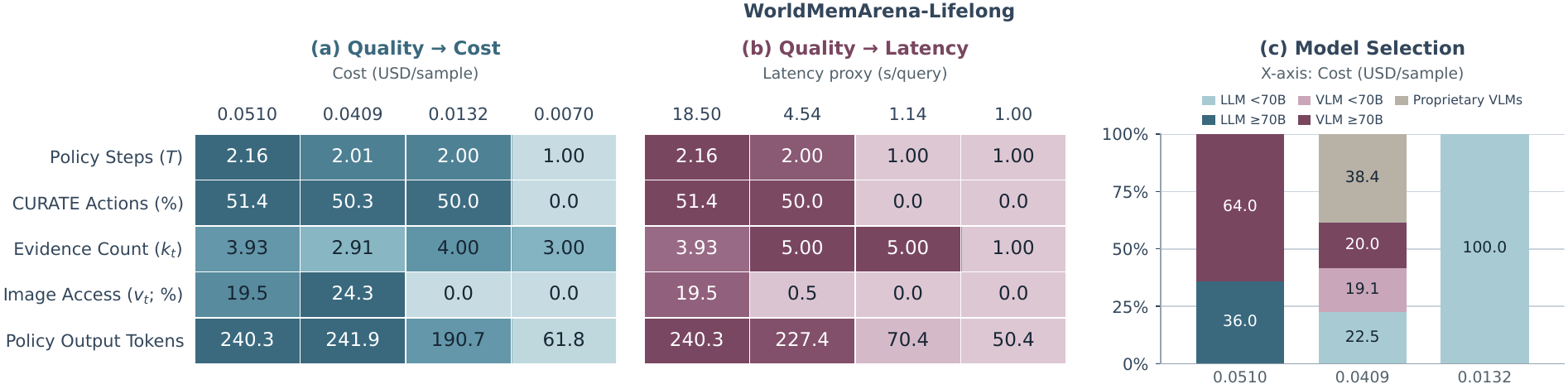}

    \vspace{2pt}

    \includegraphics[height=0.155\textheight,keepaspectratio]{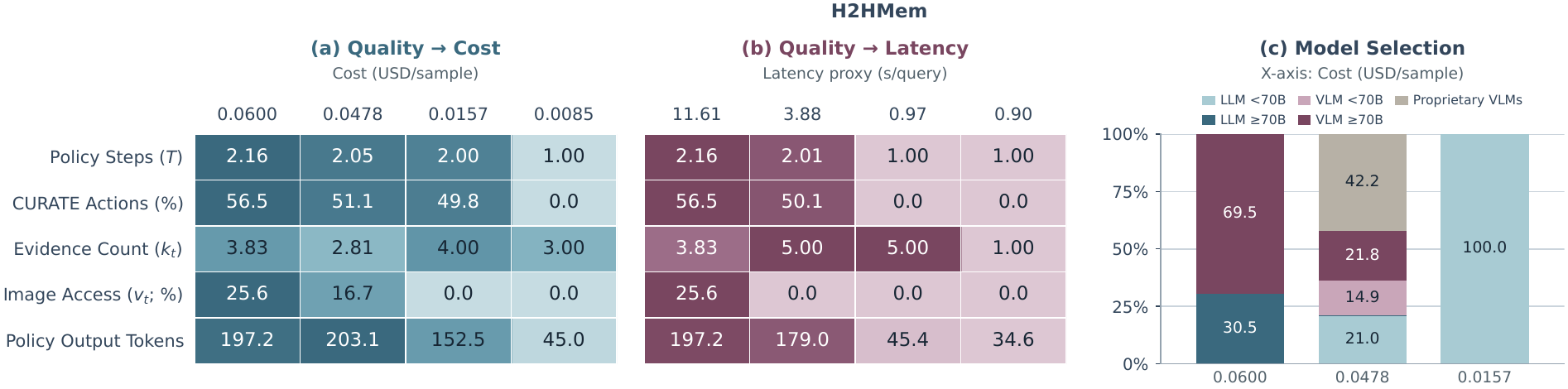}

    \vspace{2pt}

    \includegraphics[height=0.155\textheight,keepaspectratio]{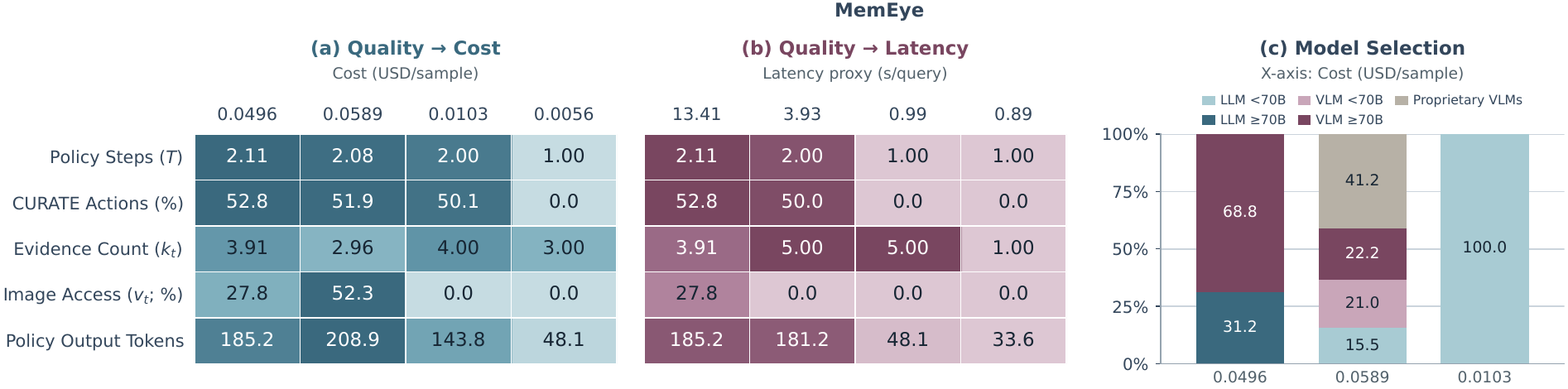}

    \vspace{2pt}

    \includegraphics[height=0.155\textheight,keepaspectratio]{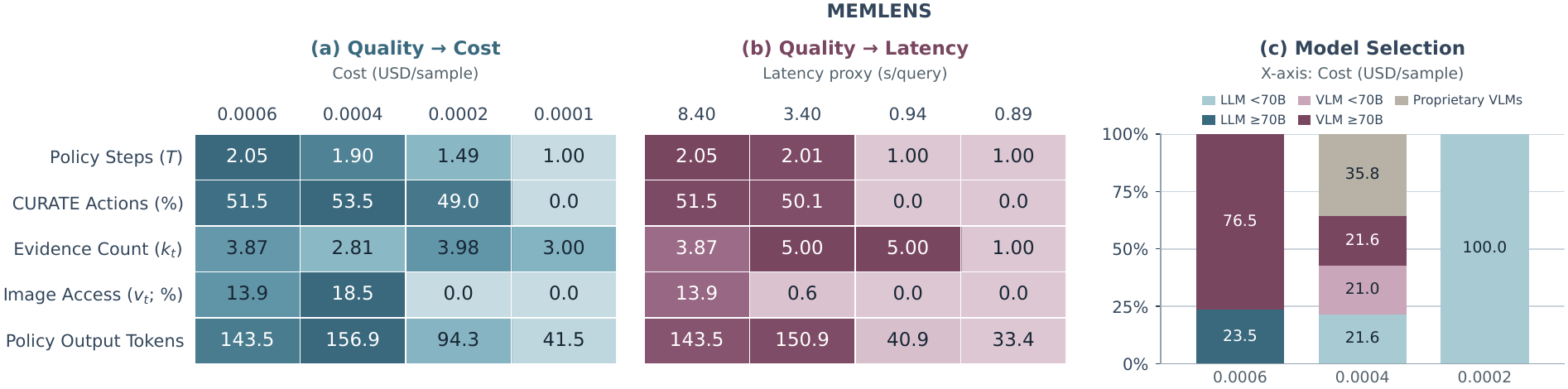}

    \caption{\textbf{Per-benchmark runtime computation allocation.}
    Runtime curation behaviors under varying cost and latency preferences on Mem-Gallery, WorldMemArena, H2HMem, MemEye, and MemLens, respectively from top to bottom.}
    \label{fig:policy_controls_per_benchmark}
\end{figure*}

\subsection{Per-Benchmark Runtime Computation Allocation}
\label{app:behavior}

Figure~\ref{fig:policy_controls_per_benchmark} provides the per-benchmark breakdown of the runtime-allocation analysis in Section~\ref{exp:behavior}. The overall behavior is consistent across all five benchmarks. As stronger cost or latency preferences are imposed, MemPilot generally reduces policy steps, \Curate{} actions, visual access, and policy-generation tokens, indicating coordinated adaptation across multiple control dimensions. 

Model routing exhibits an equally consistent shift. High-resource settings rely heavily on larger LLMs and VLMs, whereas tighter cost preferences progressively move curation toward lighter models. At the lowest displayed cost point, all five benchmarks route exclusively to LLMs below 70B. Meanwhile, individual controls such as evidence depth and image access need not change monotonically, showing that MemPilot reallocates computation jointly rather than uniformly shrinking every decision dimension. These results confirm that the aggregate behavior observed in the main text is broadly consistent across datasets while retaining dataset-specific curation patterns.

\subsection{Per-Benchmark Ablation Results}
\label{app:ablation}

Table~\ref{tab:ablation_per_benchmark} provides the per-benchmark breakdown of the ablation study. The full MemPilot configuration consistently achieves the highest Judge score across all five benchmarks. Among the ablations, removing objective-wise advantage decoupling generally preserves the strongest performance, yet still causes clear degradation on every dataset, confirming its role in multi-objective policy optimization. Removing LLM/VLM delegation, marginal utility, or retrieval--instruction decoupling leads to substantially larger drops, with similar trends observed on both in-domain and out-of-domain benchmarks. Although all ablations reduce inference cost, the accompanying quality degradation remains pronounced, indicating that the complete design yields a more effective quality--cost trade-off rather than merely consuming additional computation.

\begin{table*}[h]
    \centering
    \small
    \setlength{\tabcolsep}{4.5pt}
    \renewcommand{\arraystretch}{1.12}

    \caption{\textbf{Per-benchmark ablation results.}
    Absolute Judge scores and inference costs are reported for each variant.}
    \label{tab:ablation_per_benchmark}

    \resizebox{\textwidth}{!}{
    \begin{tabular}{l*{5}{cc}}
        \toprule
        \multirow{2}{*}{Variant}
        & \multicolumn{2}{c}{Mem-Gallery}
        & \multicolumn{2}{c}{WorldMemArena}
        & \multicolumn{2}{c}{H2HMem}
        & \multicolumn{2}{c}{MemLens$^{\dagger}$}
        & \multicolumn{2}{c}{MemEye$^{\dagger}$} \\
        \cmidrule(lr){2-3}
        \cmidrule(lr){4-5}
        \cmidrule(lr){6-7}
        \cmidrule(lr){8-9}
        \cmidrule(lr){10-11}

        & L-J $\uparrow$ & Cost $\downarrow$
        & L-J $\uparrow$ & Cost $\downarrow$
        & L-J $\uparrow$ & Cost $\downarrow$
        & L-J $\uparrow$ & Cost $\downarrow$
        & L-J $\uparrow$ & Cost $\downarrow$ \\
        \midrule

        \textbf{MemPilot-Bal}
        & \textbf{64.27} & 1.4e-2
        & \textbf{50.00} & 1.3e-2
        & \textbf{28.96} & 1.6e-2
        & \textbf{21.39} & 1.8e-4
        & \textbf{22.37} & 1.0e-2 \\
        \midrule

        w/o LLM/VLM Delegation
        & 32.00 & 8.1e-3
        & 9.09  & 6.6e-3
        & 12.08 & 8.2e-3
        & 6.94  & 1.1e-4
        & 12.80 & 5.2e-3 \\

        w/o Marginal Utility
        & 31.18 & 7.6e-3
        & 9.77  & 6.5e-3
        & 17.57 & 7.9e-3
        & 3.47  & 1.1e-4
        & 12.13 & 5.1e-3 \\

        w/o GDPO
        & 41.91 & 7.8e-3
        & 21.36 & 7.3e-3
        & 20.69 & 8.3e-3
        & 9.83  & 1.1e-4
        & 15.90 & 5.6e-3 \\

        \shortstack[l]{w/o Retrieval--Instruction\\Decoupling}
        & 34.45 & 7.5e-3
        & 10.91 & 6.4e-3
        & 16.04 & 7.8e-3
        & 3.47  & 1.1e-4
        & 11.86 & 5.0e-3 \\

        \bottomrule
    \end{tabular}
    }

    \vspace{2pt}
    \begin{minipage}{\textwidth}
    \raggedright
    \footnotesize
    $^{\dagger}$ denotes out-of-distribution (OOD) evaluation.
    \end{minipage}
\end{table*}

\subsection{Memory-Bank Compatibility Across Benchmarks}
\label{app:memory_bank}

\begin{figure*}[h]
    \centering

    \begin{subfigure}[h]{0.48\textwidth}
        \centering
        \includegraphics[width=\linewidth]{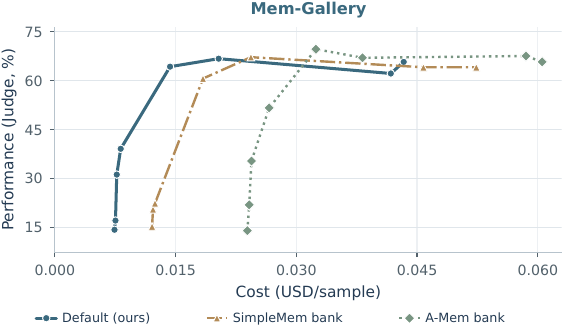}
        \caption{Mem-Gallery.}
        \label{fig:mb_memgallery}
    \end{subfigure}
    \hfill
    \begin{subfigure}[h]{0.48\textwidth}
        \centering
        \includegraphics[width=\linewidth]{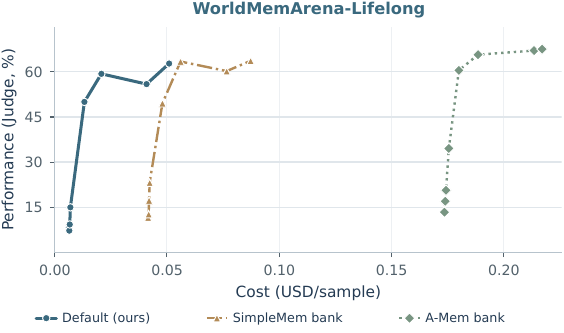}
        \caption{WorldMemArena.}
        \label{fig:mb_worldmemarena}
    \end{subfigure}

    \vspace{4pt}

    \begin{subfigure}[h]{0.48\textwidth}
        \centering
        \includegraphics[width=\linewidth]{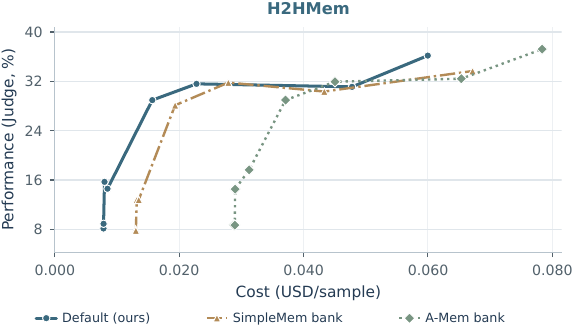}
        \caption{H2HMem.}
        \label{fig:mb_h2hmem}
    \end{subfigure}
    \hfill
    \begin{subfigure}[h]{0.48\textwidth}
        \centering
        \includegraphics[width=\linewidth]{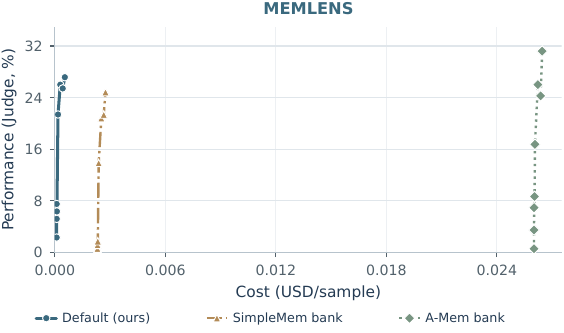}
        \caption{MemLens.}
        \label{fig:mb_memlens}
    \end{subfigure}

    \vspace{4pt}

    \begin{subfigure}[h]{0.48\textwidth}
        \centering
        \includegraphics[width=\linewidth]{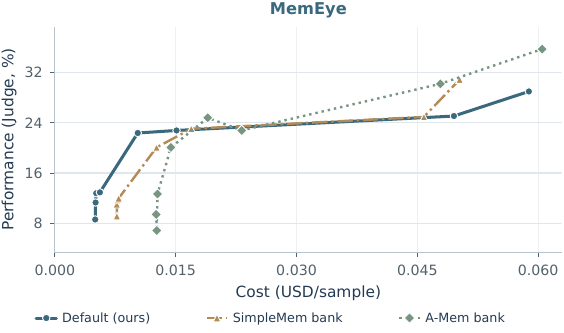}
        \caption{MemEye.}
        \label{fig:mb_memeye}
    \end{subfigure}

    \caption{\textbf{Per-benchmark memory-bank compatibility.}
    Performance--cost frontiers of MemPilot with different memory-bank initializations without retraining across five benchmarks.}
    \label{fig:memory_bank_all}
\end{figure*}

We further report per-benchmark results for the memory-bank compatibility analysis in Figure~\ref{fig:memory_bank_all}. Across all five benchmarks, replacing the default memory bank with SimpleMem~\citep{liu2026simplemem} or A-MEM~\citep{xu2025mem} preserves clear performance--cost frontiers without retraining, confirming that MemPilot is not tightly coupled to a particular query-agnostic memory construction. Different memory banks occupy distinct cost regimes while retaining similar improvements as runtime computation increases. The default bank generally provides more cost-efficient operating points, whereas A-MEM incurs higher overall cost but often reaches a higher performance ceiling, particularly on WorldMemArena, MemEye, and MemLens. SimpleMem typically lies between these regimes and remains competitive on Mem-Gallery and H2HMem. These results suggest that MemPilot preserves its runtime orchestration behavior across diverse memory banks while retaining the distinct performance--cost characteristics of their upstream memory representations.

\clearpage

\section{Case Study}
\label{app:case_study}

\begin{tcolorbox}[promptboxgray,
    title={Case 1: Direct Retrieval without Delegation}]

\textbf{Question.}
Which image shows a picnic mat? Only return the image\_id of the image.

\textbf{Step 1: Retrieve compressed memory.}
\texttt{retrieve\_memory}\\
\texttt{retrieval\_query}: image of a picnic mat\\
\texttt{evidence\_count}: 1

\textbf{Tool response (excerpt).}
\texttt{image\_id}: \texttt{D13:IMG\_001}\\
\texttt{image\_caption}: red white cloth background resembling
picnic blanket tablecloth square pattern curved edges\\
\textit{[Other retrieved text omitted.]}

\textbf{Final answer.}
\texttt{<answer>D13:IMG\_001</answer>}

\textbf{Reference answer.}
\texttt{D13:IMG\_001}

\end{tcolorbox}

\begin{tcolorbox}[promptboxgray,
    title={Case 2: Text-Only Delegation for Counting}]

\textbf{Question.}
How many Button-Down Chambray Dress options did Lin Wei show?
Please answer using an Arabic numeral
(e.g., ``2'' instead of ``Two'').

\textbf{Step 1: Retrieve compressed memory.}
\texttt{retrieve\_memory}\\
\texttt{retrieval\_query}: Button-Down Chambray Dress shown by Lin Wei\\
\texttt{evidence\_count}: 3

\textbf{Tool response (abridged).}
The returned compressed memory references dress images
\texttt{D8:IMG\_001}, \texttt{D8:IMG\_002}, and
\texttt{D8:IMG\_005}, alongside discussion of the dress options.\\
\textit{[Full compressed-memory response omitted.]}

\textbf{Step 2: Delegate raw-memory processing.}
\texttt{runtime\_memory}\\
\texttt{retrieval\_query}: Button-Down Chambray Dress shown by Lin Wei\\
\texttt{instruction}: Count the number of Button-Down Chambray
Dress options shown by Lin Wei in the retrieved memory.\\
\texttt{model}: \texttt{model\_7ab3} (3B text-only LLM)\\
\texttt{evidence\_count}: 3\\
\texttt{include\_images}: \texttt{false}

\textbf{Tool response (abridged).}
The external model enumerates five dress options, corresponding to
\texttt{D8:IMG\_001}, \texttt{D8:IMG\_002},
\texttt{D8:IMG\_003}, \texttt{D8:IMG\_004}, and
\texttt{D8:IMG\_005}.\\
\textit{[Descriptions of individual dresses omitted.]}

\textbf{Final answer.}
\texttt{<answer>5</answer>}

\textbf{Reference answer.}
5

\end{tcolorbox}

\begin{tcolorbox}[promptboxgray,
    title={Case 3: Expanded Retrieval with Image-Enabled Delegation}]

\textbf{Question.}
Which Butterfly Sleeve Sheath Dress did Lin Wei decide to buy.
Only return the image\_id of the image.

\textbf{Step 1: Retrieve compressed memory.}
\texttt{retrieve\_memory}\\
\texttt{retrieval\_query}: Butterfly Sleeve Sheath Dress purchased
by Lin Wei\\
\texttt{evidence\_count}: 1

\textbf{Tool response (excerpt).}
D8 : 16\\
: winners : chambray \# 3 butterfly sleeve \# 2 shopping detour\\
\textit{[Other retrieved text omitted; this response contains
no image ID for the selected dress.]}

\textbf{Step 2: Expand retrieval and enable image input.}
\texttt{runtime\_memory}\\
\texttt{retrieval\_query}: Butterfly Sleeve Sheath Dress purchased
by Lin Wei\\
\texttt{instruction}: Identify the image\_id of the Butterfly
Sleeve Sheath Dress that Lin Wei decided to buy based on the
provided memory content.\\
\texttt{model}: \texttt{model\_9a74}
(VLM; 235B total, 22B active)\\
\texttt{evidence\_count}: 5\\
\texttt{include\_images}: \texttt{true}

\textbf{Tool response (abridged).}
The external model associates the first dress with
\texttt{D8:IMG\_006} and the second, dark dress with red abstract
patterns with \texttt{D8:IMG\_007}. It connects Lin Wei's choice
of ``butterfly sleeve \#2'' to \texttt{D8:IMG\_007}.\\
\textit{[Remaining external-model response omitted.]}

\textbf{Final answer.}
\texttt{<answer>D8:IMG\_007</answer>}

\textbf{Reference answer.}
\texttt{D8:IMG\_007}

\end{tcolorbox}

\section{Limitations and Future Work}
\label{app:limitations-future-work}

While MemPilot provides flexible runtime memory orchestration, several aspects could be further improved:
\begin{itemize}
    \item \textbf{Privacy-aware delegation.} Runtime curation may send retrieved interaction history to external models, which can raise privacy concerns for sensitive memories. Future work could incorporate privacy-aware routing, local processing, or access-control mechanisms while remaining orthogonal to MemPilot's core orchestration design.

    \item \textbf{Preference calibration.} MemPilot controls trade-offs through predefined optimization preferences. Future work could study more user-friendly ways to map high-level deployment preferences to reward coefficients, making trade-off specification easier across applications.

    \item \textbf{Deployment-aware latency modeling.} We use a proxy latency estimator during training to avoid instability from transient network and serving fluctuations. Future work could instead optimize directly against measured latency in a controlled system-level sandbox with stable networking and serving conditions.
\end{itemize}

\section{Broader Impact}
\label{app:broader-impact}

MemPilot advocates a broader shift in agent memory from performance-oriented optimization toward explicit and controllable performance--cost--latency trade-offs. In practical deployments, the highest-performing memory configuration is not always the most desirable, as different users and applications may place substantially different values on monetary cost and response latency. This mirrors modern commercial LLM services, which increasingly offer model tiers spanning different capability, cost, and latency profiles (e.g., GPT-5.6-Sol/Terra/Luna), allowing users to choose operating points that best match their needs. We believe agent memory systems should provide similar flexibility rather than committing to a single performance-oriented configuration. This becomes particularly relevant for multimodal agents, where accessing and processing visual evidence can introduce substantially different resource demands. More broadly, treating memory curation as adaptive computation allocation may encourage tighter integration between agent memory, model routing, and resource-aware decision making, enabling future long-horizon agents to become more efficient, responsive, and deployment-aware.

\section{Prompts}
\label{app:prompt-display-templates}

\subsection{Policy Prompt}
\label{app:policy-prompt}

The policy shares the following instructions across benchmarks.
Benchmark-specific inputs and answer requirements are represented
by placeholders. Tool schemas specify valid arguments and model IDs.
The tools \texttt{retrieve\_memory} and \texttt{runtime\_memory}
implement Retrieve and CURATE, respectively.

\begin{tcolorbox}[promptbox, title={Policy Prompt Template}]

\textbf{Role.}
You are a runtime-memory agent that solves questions by planning and
executing an adaptive memory pipeline, directly retrieving compressed
evidence or delegating raw memory chunks to capability-diverse external
LLMs for processing as needed.

\textbf{Input (user message).}
\texttt{\detokenize{<question>}}, with
\texttt{\detokenize{<benchmark-specific metadata and answer constraints>}}
when applicable.

\textbf{Instruction.}
Work iteratively:
\begin{itemize}
    \item Use \texttt{retrieve\_memory} for faster access to precomputed
    query-agnostic compressed memory, which may omit important details.
    Use \texttt{runtime\_memory} to delegate query-aware processing to a
    selected external LLM using the complete, uncompressed content of
    each retrieved raw memory chunk, while incurring some additional
    model-call cost and latency.

    \item For visual content, \texttt{retrieve\_memory} exposes only
    textual captions and cannot inspect image pixels. To inspect pixels
    from retrieved or question images, use \texttt{runtime\_memory}
    with an image-capable LLM and set \texttt{include\_images}
    to \texttt{true}.

    \item Before the first call, decompose the query into one or more
    atomic, evidence-dependent subtasks. Execute only the next
    unresolved subtask; do not delegate multiple dependent subtasks
    through one broad instruction.

    \item Before each tool call, briefly assess what is known, what
    remains unresolved, and why the next call is useful.
    After each memory-tool result, update this assessment before
    deciding on another call or the final answer.

    \item Execute dependent subtasks sequentially. Calls in the same
    assistant message run in parallel, so put only independent subtasks
    in separate \texttt{tool\_call} blocks, at most two per message.
    If a later call depends on an earlier memory-tool result,
    issue it in a later turn.
\end{itemize}

\textcolor{pppromptamber}{\textbf{Evidence constraint.}}
Each successful tool result contains findings supported by that call's
inputs and may identify unresolved parts. Missing evidence is local to
the selected inputs, not proof that the full query is unanswerable.
Use partial findings to plan later calls or the final answer.

\textbf{Tool-call format.}
Available call forms (serialization only; neither is preferred):

\emph{Direct compressed-memory retrieval:}
\begin{flushleft}
\small\ttfamily
\detokenize{<tool_call>}\\
\{"name": "retrieve\_memory",\\
\hspace*{1em}"arguments": \{\\
\hspace*{2em}"retrieval\_query":\\
\hspace*{3em}"<self-contained evidence query>",\\
\hspace*{2em}"evidence\_count": <integer>\\
\hspace*{1em}\}\}\\
\detokenize{</tool_call>}
\end{flushleft}

\emph{Routed original-memory processing:}
\begin{flushleft}
\small\ttfamily
\detokenize{<tool_call>}\\
\{"name": "runtime\_memory",\\
\hspace*{1em}"arguments": \{\\
\hspace*{2em}"retrieval\_query":\\
\hspace*{3em}"<self-contained evidence query>",\\
\hspace*{2em}"instruction":\\
\hspace*{3em}"<evidence-processing instruction>",\\
\hspace*{2em}"model": "<model ID>",\\
\hspace*{2em}"evidence\_count": <integer>,\\
\hspace*{2em}"include\_images": true\\
\hspace*{1em}\}\}\\
\detokenize{</tool_call>}
\end{flushleft}

Choose every argument value for the current step.

\textbf{Output.}
Once you have enough evidence, stop calling tools and finish with
\texttt{\detokenize{<benchmark-specific final-answer contract>}}.
Do not write anything after the closing
\texttt{\detokenize{</answer>}} tag.

\end{tcolorbox}

\subsection{Delegation Prompt}
\label{app:delegation-prompt}

The selected LLM/VLM receives the policy-generated instruction
and retrieved raw evidence. The retrieval query is used for search
and is not separately included in this prompt.
Image-related sections are optional; image pixels are attached
when requested and supported by the selected model.

\begin{tcolorbox}[promptboxblue, title={Delegation Prompt Template}]

\textbf{System.}
You are a memory-processing assistant. Follow the instruction using
only the question-image descriptions, memory items, and images in
the input. Return all findings supported by these inputs.
If the instruction can only be partially completed, return the
supported findings and briefly state what remains unresolved.
If nothing relevant is present, say so directly.
Do not add unsupported information.

\medskip
\textbf{User.}

\textbf{Instruction:}\\
\texttt{\detokenize{<policy-generated evidence-processing instruction>}}

\textbf{Question image descriptions:}\\
\texttt{\detokenize{<question-image IDs and captions, if present>}}

\textbf{Memory items:}\\
\texttt{\detokenize{<retrieved raw memory chunks with IDs and metadata>}}

\textbf{Additional image inputs:}\\
\texttt{\detokenize{<selected memory-image IDs and captions, if present>}}

\end{tcolorbox}

\subsection{Question-Answering Prompt}
\label{app:qa-prompt}

The answer model receives the question and memory evidence returned
by the policy, without the policy's reasoning or original
final answer. The shared template uses the benchmark-specific
instructions listed below.

\begin{tcolorbox}[promptboxgreen, title={QA Prompt Template}]

\textbf{System.}
\texttt{\detokenize{<benchmark-specific QA instructions>}}

Return only one non-empty
\texttt{\detokenize{<answer>...</answer>}}
block, with the answer text inside the tags.

\medskip
\textbf{User.}

\textbf{\texttt{\detokenize{<memory heading>}}:}\\
\texttt{[Evidence 1]}\\
\texttt{\detokenize{<memory evidence returned by a policy tool>}}

\texttt{[Evidence 2]}\\
\texttt{\detokenize{<additional memory evidence>}}

\texttt{\ldots}

\textbf{Question Image:}\\
\texttt{\detokenize{<question-image IDs and captions, if available>}}

\textbf{Question Date:}\\
\texttt{\detokenize{<question date, if provided for MemLens>}}

\textbf{Question:}\\
\texttt{\detokenize{<question>}}

\texttt{\detokenize{<question-specific output constraint, if applicable>}}

\end{tcolorbox}

The following instructions replace the benchmark-specific
placeholder in the system message.

\begin{tcolorbox}[promptboxgreen,
    title={QA Instructions: Mem-Gallery}]

Answer from the provided multimodal conversation memory.
Ground the answer in that memory, prefer the latest information
when details change over time, and keep the response concise
but complete. When the question asks for images, preserve the
exact \texttt{image\_id} values.

\medskip
\textit{Additional user-message constraint for conflict detection:}

The text inside
\texttt{\detokenize{<answer>...</answer>}}
must be exactly ``Yes.'' or ``No.''

\medskip
\textit{Additional user-message constraint for visual search:}

Return the exact matching \texttt{image\_id} value or values.
If several apply, sort them in ascending order and separate
them with commas.

\end{tcolorbox}

\begin{tcolorbox}[promptboxgreen,
    title={QA Instructions: WorldMemArena}]

Answer from the provided retrieved memories.
Synthesize multiple entries when needed, use timestamps and the
most recent evidence to resolve conflicts, and ground every
claim in the memories. Convert relative dates using the memory
timestamps. When images are requested, preserve exact
\texttt{image\_id} values.
Keep the answer concise; if no relevant information is present,
the answer text must be exactly ``Not mentioned in memory.''

\end{tcolorbox}

\begin{tcolorbox}[promptboxgreen,
    title={QA Instructions: H2HMem}]

You are a memory testing system.
\texttt{\detokenize{<question-type instruction>}}
Answer directly in English without reasoning, using no more
than 100 words.

\medskip
\textit{Additional system instruction for conflict detection:}

The answer text must be strictly either Yes or No.

\medskip
\textit{Additional system instruction for answer refusal:}

If the information is absent, the answer text must be exactly
``Not mentioned.''

\end{tcolorbox}

For H2HMem, the question-type instruction is selected from
the following:

\begin{tcolorbox}[promptboxgreen,
    title={H2HMem Question-Type Instructions}]

\textbf{Unimodal precise recall.}
Accurately recall the requested information from the
conversation memory.

\textbf{Cross-modal related retrieval.}
Use related textual and visual information from the
conversation memory.

\textbf{Knowledge resolution.}
Resolve knowledge consistently across the conversation memory.

\textbf{Temporal reasoning.}
Reason about temporal relationships in the conversation memory.

\textbf{Multimodal causal reasoning.}
Use textual and visual evidence for the requested causal reasoning.

\textbf{Reference \& evolution tracking.}
Track references and how they evolve across the conversation memory.

\textbf{Test-time learning.}
Use knowledge introduced in the conversation memory to answer
the question.

\textbf{Conflict detection.}
Determine whether the information in the question conflicts
with the conversation memory.

\textbf{Answer refusal.}
Determine whether the question can be answered from the
conversation memory.

\end{tcolorbox}

\begin{tcolorbox}[promptboxgreen,
    title={QA Instructions: MemEye (Open-Ended)}]

Answer from the provided long-horizon multimodal conversation
memory. Ground the answer in the memory and relevant visual
evidence; when evidence conflicts, use the newest visual evidence
unless the question itself asks about the conflict.
For counts, give the number; for yes/no questions, begin with
Yes or No; keep descriptive answers to one or two sentences.
If the requested information is absent, state that it is
unavailable.

\end{tcolorbox}

\begin{tcolorbox}[promptboxgreen,
    title={QA Instructions: MemLens}]

Answer from the provided conversation history.
If the question cannot be answered from it, the answer text
must be exactly ``Insufficient information''.

\end{tcolorbox}

\subsection{Answer-Judging Prompts}
\label{app:judge-prompt}

\paragraph{Mem-Gallery and H2HMem.}

These benchmarks share the following five-level rubric.
In the output-format example, the reasoning placeholder is empty
for Mem-Gallery and \texttt{\detokenize{<short explanation>}}
for H2HMem.

\begin{tcolorbox}[
    promptboxred,
    breakable,
    title={Judge Prompt: Mem-Gallery and H2HMem}
]

You are an impartial judge evaluating the memory capabilities
of an AI assistant with the question-answering task.

Your task is to compare the Assistant's Answer against the
Ground Truth and assign a score of 0, 0.25, 0.5, 0.75, or 1.

\textbf{Scoring Rubric.}

\textbf{Score 0 (Incorrect / Miss):}
\begin{itemize}
    \item The answer contradicts the Ground Truth.
    \item For Yes/No questions: The answer has the wrong polarity
    (e.g., says ``Yes'' when Ground Truth is ``No'').
    \item For Open-ended questions: The answer provides factually
    wrong information or hallucinations.
    \item The assistant fails to provide the required information.
\end{itemize}

\textbf{Score 0.25 (Poor / Tangential):}
\begin{itemize}
    \item The answer touches on the topic but misses the
    \textbf{core entity} or key value required.
    \item The answer contains a mix of minor correct details and
    \textbf{significant hallucinations} or wrong associations.
    \item The answer is excessively vague to the point of being
    useless (e.g., answering ``a dog'' instead of
    ``a golden retriever'').
\end{itemize}

\textbf{Score 0.5 (Partial / Vague):}
\begin{itemize}
    \item The answer is technically correct, but lacks confidence
    or is incomplete.
    \item The answer captures the \textbf{main entity or concept}
    correctly but misses a part of the required supporting details.
    \item For Yes/No questions: The polarity is correct, but the
    reasoning is flawed (if have), or the assistant is uncertain
    (e.g., ``I think it might be Yes'').
    \item For Open-ended questions: The answer is too general or
    misses key adjectives/details present in the Ground Truth.
\end{itemize}

\textbf{Score 0.75 (Good / Minor Imperfection):}
\begin{itemize}
    \item The answer is largely accurate and captures the core
    information confidently.
    \item It misses only \textbf{minor details}
    (e.g., specific adjectives or secondary details)
    that do not alter the main truth.
    \item The answer contains all the correct information but
    includes unnecessary ``fluff'' or slight conversational filler
    that reduces precision.
\end{itemize}

\textbf{Score 1 (Correct / Exact):}
\begin{itemize}
    \item The answer is accurate, precise, and confident.
    \item For Yes/No questions: The polarity matches the
    Ground Truth perfectly.
    \item For Open-ended questions: The answer contains
    \textbf{all} the core information and necessary details
    required by the Ground Truth without hallucinations.
\end{itemize}

\textbf{Input Data.}

Ground Truth:
\texttt{\detokenize{<reference answer(s)>}}

Assistant Answer:
\texttt{\detokenize{<predicted answer>}}

\textbf{Output Format.}

Output strictly in the following JSON format:
\begin{flushleft}
\small\ttfamily
\{"score": <0, 0.25, 0.5, 0.75, or 1>,\par
\hspace*{1em}"reasoning": "<reasoning placeholder>"\}
\end{flushleft}

\end{tcolorbox}

\paragraph{WorldMemArena.}

The evaluator assigns a score of 1 to \texttt{Correct}
and 0 to \texttt{Hallucination} or \texttt{Omission}.
This evaluates answer correctness, not evidence coverage.

\begin{tcolorbox}[
    promptboxred,
    breakable,
    title={Judge Prompt: WorldMemArena}
]

You are an \textbf{evaluation expert for AI memory system
question answering}.

Based \textbf{only} on the provided \textbf{Reference Answer},
strictly evaluate the \textbf{accuracy} of the
\textbf{Memory System Response}.
Classify it as one of \textbf{Correct},
\textbf{Hallucination}, or \textbf{Omission}.
Do \textbf{not} use any external knowledge or subjective inference.

\textbf{Evaluation Criteria.}

\textbf{1. Correct}
\begin{itemize}
    \item The response accurately answers the question and is
    \textbf{semantically equivalent} to the Reference Answer.
    \item No contradictions with the Reference Answer.
    \item Synonyms, paraphrasing, and reasonable summarization
    are acceptable.
\end{itemize}

\textbf{2. Hallucination}
\begin{itemize}
    \item The response includes information that
    \textbf{contradicts} the Reference Answer.
    \item When the Reference Answer is \emph{unknown/uncertain},
    yet the response provides a specific fact.
\end{itemize}

\textbf{3. Omission}
\begin{itemize}
    \item The response is \textbf{incomplete} compared to the
    Reference Answer.
    \item It states ``don't know'' or ``no related memory''
    even though the Reference Answer supplies the answer.
    \item For multi-element answers, missing \textbf{any}
    element counts as Omission.
\end{itemize}

\textbf{Priority Rules.}
\begin{itemize}
    \item Both missing info AND fabricated info
    \(\rightarrow\) \textbf{Hallucination}.
    \item No fabrication but missing info
    \(\rightarrow\) \textbf{Omission}.
    \item Fully equivalent
    \(\rightarrow\) \textbf{Correct}.
\end{itemize}

\textbf{Information.}

Reference Answer:
\texttt{\detokenize{<reference answer(s)>}}

Memory System Response:
\texttt{\detokenize{<predicted answer>}}

\textbf{Output.}
\begin{flushleft}
\small\ttfamily
\{\par
\hspace*{1em}"reasoning": "Concise evaluation rationale",\par
\hspace*{1em}"evaluation\_result":\par
\hspace*{2em}"Correct | Hallucination | Omission"\par
\}
\end{flushleft}

\end{tcolorbox}

\paragraph{MemEye.}

The judge produces a five-level raw score.
The evaluator subsequently maps raw scores
\(0, 0.25, 0.5, 0.75, 1\) to
\(0, 0.5, 0.5, 1, 1\), respectively.
This mapping is post-processing and is not part of the prompt.

\begin{tcolorbox}[
    promptboxred,
    breakable,
    title={Judge Prompt: MemEye (Open-Ended)}
]

You are an impartial judge evaluating the memory capabilities
of an AI assistant on a question-answering task.

Your task is to compare the Assistant's Answer against the
Ground Truth and assign a score of 0, 0.25, 0.5, 0.75, or 1.

\textbf{Important Principles.}

\begin{enumerate}
    \item \textbf{Semantic equivalence over surface form.}
    The assistant may use different wording, synonyms, or sentence
    structure. As long as the meaning matches the Ground Truth,
    it should receive full credit.
    Do NOT penalize for stylistic differences.

    \item \textbf{Numeric / counting questions are binary.}
    If the Ground Truth is a specific number (e.g., ``3''),
    the answer is either exactly right (score 1) or wrong (score 0).
    There is no partial credit for being off by one.

    \item \textbf{Negation + correction questions.}
    Some Ground Truth answers first negate a premise and then
    provide the correct information
    (e.g., ``No --- it was May 18-22, not May 15-19'').
    The assistant must get BOTH the negation AND the corrected
    fact right for full credit.

    \item \textbf{Identity / ``Who is'' questions.}
    The Ground Truth may describe a person by role, actions,
    or relationships rather than by name.
    The assistant's answer is correct if it identifies the same
    unique person, even if it uses a name instead of a description
    or vice versa.

    \item \textbf{Answer the thing that was asked.}
    Judge whether the assistant answered the requested attribute
    or type of information. If the question asks for a category,
    role, relationship, time, count, ordering, or specific item,
    an answer giving a different type of information is not fully
    correct even if it is related.

    \item \textbf{Wrong entity beats related entity.}
    If the answer names the wrong person, object, brand, place,
    event, or option, do not give high scores just because it
    is on-topic. Being from the right semantic domain is not enough.

    \item \textbf{Multi-item questions require set matching.}
    Full credit requires all required items and no incorrect extras.
    If only some required items are correct, partial credit
    may be appropriate.

    \item \textbf{Ordering / chronology questions require
    the correct item.}
    If the question asks which item was first, last, middle,
    earlier, later, before, or after, the selected item must
    match the Ground Truth.
    A different plausible item is still wrong.

    \item \textbf{Do not over-reward fluent explanations.}
    Extra detail is helpful only if it stays consistent with
    the Ground Truth.
    Confident wording or elaboration should not increase the
    score when the core fact is wrong.

    \item \textbf{Contradicted relations or orderings
    are usually zero.}
    If the question asks about a relationship such as before/after,
    earlier/later, first/last, middle, sequence, relative timing,
    or other ordering, and the assistant gives the opposite
    relation, denies the relation, or selects the wrong ordered
    item, this is usually a clear contradiction and should
    normally receive score 0 rather than 0.25.
\end{enumerate}

\textbf{Scoring Rubric.}

\textbf{Score 0 (Incorrect / Miss):}
\begin{itemize}
    \item The answer contradicts the Ground Truth.
    \item For Yes/No questions: The answer has the wrong polarity
    (e.g., says ``Yes'' when Ground Truth is ``No'').
    \item For numeric questions: The number is wrong.
    \item For attribute-type mismatches: the answer gives the
    wrong kind of thing (for example, a brand when the question
    asks for a product category, or a related event when the
    question asks for the earliest specific item).
    \item For relation/order questions: if the answer contradicts
    the required before/after, earlier/later, first/last, middle,
    or similar relationship, score 0 in the usual case.
    \item The answer provides factually wrong information,
    hallucinations, or fails to provide any relevant information.
\end{itemize}

\textbf{Score 0.25 (Poor / Tangential):}
\begin{itemize}
    \item The answer touches on the topic but misses the
    \textbf{core entity} or key value required.
    \item The answer contains a mix of minor correct details and
    \textbf{significant hallucinations} or wrong associations.
    \item The answer is excessively vague to the point of being
    useless (e.g., answering ``a dog'' instead of
    ``a golden retriever'').
    \item For multi-item questions: use 0.25 when the answer is
    clearly about the right topic but most of the required
    set is wrong.
\end{itemize}

\textbf{Score 0.5 (Partial / Vague):}
\begin{itemize}
    \item The answer captures the \textbf{main entity or concept}
    correctly but misses important supporting details.
    \item For negation + correction questions: The polarity is
    correct but the corrected fact is missing or wrong,
    or vice versa.
    \item The answer is too general or misses key qualifiers
    present in the Ground Truth.
    \item For multi-item questions: use 0.5 when the answer gets
    a substantial part of the required set right
    (for example, one of two required items is correct and the
    rest is wrong or missing).
\end{itemize}

\textbf{Score 0.75 (Good / Minor Imperfection):}
\begin{itemize}
    \item The answer is largely accurate and captures the
    core information confidently.
    \item It misses only \textbf{minor details}
    (e.g., specific adjectives or secondary details)
    that do not alter the main truth.
    \item The answer contains all the correct information but
    includes unnecessary filler that reduces precision.
\end{itemize}

\textbf{Score 1 (Correct / Exact):}
\begin{itemize}
    \item The answer is accurate, precise, and confident.
    \item For Yes/No questions: The polarity matches the
    Ground Truth perfectly.
    \item For numeric questions: The number matches exactly.
    \item For descriptive questions: The answer contains
    \textbf{all} the core information and necessary details
    required by the Ground Truth without hallucinations.
    \item Semantic equivalence is sufficient --- exact wording
    is NOT required.
    \item For multi-item questions: all required items are
    present and there are no incorrect extra items.
\end{itemize}

\textbf{Input Data.}

Ground Truth:
\texttt{\detokenize{<reference answer(s)>}}

Assistant Answer:
\texttt{\detokenize{<predicted answer>}}

\textbf{Output Format.}

Output strictly in the following JSON format:
\begin{flushleft}
\small\ttfamily
\{"score": <0, 0.25, 0.5, 0.75, or 1>,\par
\hspace*{1em}"reasoning": "<short explanation>"\}
\end{flushleft}

\end{tcolorbox}

\paragraph{MemLens.}

MemLens uses a binary rubric with a task-specific criterion.
Exactly one of the criteria listed below is inserted into
the shared template.

\begin{tcolorbox}[
    promptboxred,
    breakable,
    title={Judge Prompt: MemLens}
]

Now your role is a grading teacher.
Your task is to review and score student answers based on
reference standard answers for a question-answering benchmark.

First, extract the final answer from the student's solution,
then judge whether the answer is correct.
Score only the final answer; intermediate problem-solving steps
do not need to be correct.
If multiple inconsistent answers are given, or no clear final
answer is stated, assign 0 points.
Only the student's last clearly stated answer counts.

An item is covered if it is strictly mentioned or unambiguously
implied by semantic equivalence.
This includes numerical equivalence, synonyms, plural/singular
forms, and equivalent date formats.
Do not accept loosely related concepts.
Ignore minor formatting differences, capitalization, punctuation,
and equivalent wording when meaning is unchanged.

The scoring scale has two levels: 1 point and 0 points.
Assign 1 point if the student's final answer matches the standard
answer under the task-specific criteria below.
Assign 0 points otherwise, including refusal or a claim of
insufficient information except for answer-refusal tasks.

\textbf{[Task-Specific Criteria]:}

\texttt{\detokenize{<criterion selected for the question type>}}

Your output format is:
\begin{flushleft}
\small\ttfamily
[Scoring Rationale]:\par
[Score]: x points\par
[JSON]: \{"answer\_score": <0 or 1>\}
\end{flushleft}

\textbf{[Current Case]}

Standard Answer:
\texttt{\detokenize{<reference answer(s)>}}

Student Answer:
\texttt{\detokenize{<predicted answer>}}

\textbf{[Scoring Rationale]:}

\end{tcolorbox}

\begin{tcolorbox}[
    promptboxred,
    breakable,
    title={MemLens Task-Specific Criteria}
]

\textbf{[IE --- Entity / Attribute Extraction]}

This is an information extraction question about a concrete
entity, count, attribute, text, object, location, or visual
detail from the provided memory context.
\begin{itemize}
    \item Assign 1 point if the student's response contains
    the core information from the standard answer.
    Minor wording differences are acceptable, but the essential
    information must be present and correct.
    \item Assign 0 points if the core information is missing,
    contradicted, too vague, refused, or incorrect.
    \item For numeric or count answers, require the exact value
    unless the standard answer itself is approximate.
\end{itemize}

\textbf{[IE --- Previous Information Extraction]}

This is an information extraction question about previously
mentioned information, spatial relations, counts, or attributes
from earlier conversation/image context.
\begin{itemize}
    \item Assign 1 point if the student's response recovers
    the same previous information as the standard answer,
    including the correct relation, location, count, or attribute.
    \item Minor wording differences are acceptable if they
    preserve the same meaning.
    \item Assign 0 points if the answer refers to the wrong
    previous item, gives the wrong relation/count/location,
    is too vague, refuses, or is unsupported.
\end{itemize}

\textbf{[KU --- Knowledge Update]}

This is a knowledge update question testing whether the model
tracks the latest state.
\begin{itemize}
    \item Assign 1 point if the student gives the current
    (correct) answer.
    \item Assign 0 points if the student gives an outdated answer,
    any other incorrect answer, refuses, or does not clearly answer.
\end{itemize}

\textbf{[MSR --- Yes/No (Entity Resolution)]}

This is a yes/no question.
\begin{itemize}
    \item Assign 1 point if the student's final answer is
    semantically equivalent to the standard answer
    (e.g., ``Yes'' matches ``Yes'', ``No'' matches ``No''),
    even if phrased differently
    (e.g., ``That's correct'' for ``Yes'').
    \item Assign 0 points if the answer means the opposite,
    refuses, or does not clearly answer.
\end{itemize}

\textbf{[MSR --- Counting]}

This is a counting question.
The standard answer is a specific number.
\begin{itemize}
    \item Assign 1 point if the student clearly states the
    exact number in any equivalent form
    (e.g., ``3'', ``three'', ``3.0'').
    \item Assign 0 points if the student gives a different number,
    a range, an approximation, refuses, or does not clearly answer.
\end{itemize}

\textbf{[MSR --- Arithmetic]}

This is an arithmetic question.
The standard answer is a specific value.
\begin{itemize}
    \item Assign 1 point if the student gives the same value
    in an equivalent format
    (e.g., ``0.5'' and ``1/2'' and ``50\%'').
    \item Assign 0 points if the student gives a different value,
    an unsupported approximation, refuses, or does not clearly answer.
\end{itemize}

\textbf{[TR --- Duration Comparison (A/B)]}

This is a binary choice question.
The student must select one of two options.
\begin{itemize}
    \item Assign 1 point if the student clearly selects the
    same option as the standard answer, regardless of phrasing.
    \item Assign 0 points if the student selects the other option,
    refuses, or does not make a clear selection.
\end{itemize}

\textbf{[TR --- Order Ranking]}

This is an ordering question.
The standard answer is an exact sequence of items.
\begin{itemize}
    \item Assign 1 point ONLY if the student gives the exact
    same sequence in the same order.
    \item Assign 0 points if any item is missing, added,
    duplicated, misplaced, or if the order is unclear.
\end{itemize}

\textbf{[TR --- Date Extraction]}

This is a date extraction question.
The standard answer is a specific date.
\begin{itemize}
    \item Assign 1 point if the student gives the same date
    in any unambiguous equivalent format
    (e.g., ``Jan 15, 2024'' and ``2024-01-15'' and ``15/01/2024'').
    \item Assign 0 points if the date is different,
    incomplete when a full date is required, ambiguous, or missing.
\end{itemize}

\textbf{[AR --- Answer Refusal]}
\begin{itemize}
    \item Assign 1 point if the student explicitly refuses,
    says there is not enough information, or indicates the
    question is unanswerable from the provided context.
    \item Assign 0 points if the student gives a substantive
    answer, even if hedged.
\end{itemize}

\end{tcolorbox}

\end{document}